\documentclass{article}

 \usepackage{natbib}
 \usepackage{url}
 \usepackage{hyperref}
\usepackage{algorithm}
\usepackage{algpseudocode}
\usepackage{pgfplots}
\pgfplotsset{compat=1.17}
\usepgfplotslibrary{groupplots}
\usepackage{amsmath,amssymb,amsthm}

\usepackage[left=1.5cm, right=2cm]{geometry}

 \newgeometry{left=3cm,bottom=4cm}

\tikzset{
  background/.style={%
    execute at begin node={\begin{pgfonlayer}{bg}},
    execute at end node={\end{pgfonlayer}}
  }
}

\title{Multi-Objective {\it min-max} Online Convex Optimization}

\usepackage{times}
\usepackage{latexsym}
\usepackage{verbatim}
\newtheorem{theorem}{Theorem}
\newtheorem{assumption}[theorem]{Assumption}
\newtheorem{remark}{Remark}
\newtheorem{lemma}[theorem]{Lemma}

\def\bb0{{\mathbb{0}}}

\def\bb{{\mathbf{b}}}

\def\b0{{\mathbf{0}}}
\def\opt{\mathsf{OPT}}

\def\b1{{\mathbf{1}}}

\def\bbE{{\mathbb{E}}}

\def\bbR{{\mathbb{R}}}

\def\cA{\mathcal{A}}
\def\cB{\mathcal{B}}

\def\cD{\mathcal{D}}

\def\cP{\mathcal{P}}

\def\cR{\mathcal{R}}
\def\cS{\mathcal{S}}

\def\cX{\mathcal{X}}

\def\sfr{{\mathsf{r}}}

\def\sf0{{\mathsf{0}}}

\def\nn{\nonumber}

\author{
    Rahul Vaze\\
    School of Technology and Computer Science\\
    Tata Institute of Fundamental Research, Mumbai\\
    rahul.vaze@gmail.com
  \and
    Sumiran Mishra\\
    Indian Institute of Science Education and Research, Pune\\
    sumiran.mishra@students.iiserpune.ac.in
}
\usepackage{enumitem}
\begin{document}

\maketitle

\begin{abstract}

In this paper, we broaden the horizon of online convex optimization (OCO), and consider multi-objective OCO, where there are $K$ distinct loss function sequences, and an algorithm has to choose its action at time $t$, before the $K$ loss functions 
at time $t$ are revealed. To capture the tradeoff between tracking the $K$ different sequences, we consider the {\it min-max} regret, where the benchmark (optimal offline algorithm) takes a static action across all time slots that minimizes the maximum of the total loss (summed across time slots) incurred by each of the $K$ sequences. An online algorithm is allowed to change its action across time slots, and its {\it min-max} regret is defined as the difference between its {\it min-max} cost and that of the benchmark.
The {\it min-max} regret is a stringent performance measure and an algorithm with small regret needs to `track' all loss functions simultaneously. 

We first show that with adversarial input, {\it min-max} regret scales linearly with the time horizon $T$ for any online algorithm.
Consequently, we consider a stochastic i.i.d. input model where all loss functions are i.i.d. generated from an unknown joint distribution and propose a simple algorithm that combines the well-known {\it Hedge} and online gradient descent (OGD) and show via a remarkably simple proof that its expected {\it min-max} regret is $O(\sqrt{T \log (T K)})$. Analogous results are also derived for Martingale difference and Markov input models. 
The binary prediction problem with multiple binary sequences is also considered where an algorithm has to predict a binary action and the loss function is 
the maximum of the Hamming distance between the predicted actions and the binary sequences. 
For this problem, we again show that under adversarial input the {\it min-max} regret grows linearly with the time horizon 
$T$ for any online algorithm, while under the stochastic i.i.d. input model we propose an algorithm that achieves 
$O(1)$ {\it min-max} regret.



\end{abstract}

\section{Introduction}
Online convex optimization (OCO) \cite{HazanBook} is a basic learning problem that models a wide variety of applications such as spam filtering \cite{HazanBook}, capacity provisioning in cloud networks \cite{lin2012dynamic} etc., where at time $t$, an online algorithm $\cA$ has 
to choose an action $x_t \in \cX$, after which the adversary reveals the convex objective/loss function $f_t$ on which $x_t$ is evaluated. As a benchmark, the optimal offline  algorithm is considered, that knows the entire set of $f_t$, $t=1, \dots, T$, however, can only choose a single action across all times. 
To quantify the performance of online algorithms with OCO, the metric of {\it static regret} $R^{\text single}_\cA(T) = \sup_{f_t}\left\{\sum_{t=1}^T f_t(x_t) - \min_x \sum_{t=1}^T f_t(x)\right\}$ is studied, 
 where $x_t$'s are the actions taken by $\cA$.


In a long line of work, comprehensively summarized in \cite{HazanBook}, tight guarantees (both upper and lower bounds) have been obtained for the OCO problem, where optimal $R^{\text single}_\cA(T)= \Theta(\sqrt{T})$. In particular, online gradient descent (OGD), mirror descent, and follow-the-regularized-leader (FTRL) \cite{HazanBook} all achieve the optimal static regret performance for OCO.

Many modern real-world decision problems, however, involve multiple conflicting performance metrics, motivating the multi-objective OCO framework—an extension of OCO that aims to achieve Pareto-efficient trade-offs among competing objectives. However, Pareto optimality is too challenging to work with in the online setting, both in terms of defining the right benchmark as well as the required analytical treatment.
One such attempt was made at this recently in \cite{jiang2023multiobjective}, however, it used a {\it weak} regret metric, which remained independent of the number of objectives. We provide more details of this in the Related Work section. 

Thus, in this paper, we consider the robust alternative in the multi-objective case: the {\it min-max} OCO, where the objective of an algorithm is to minimize the maximum of the total loss (summed across time slots) incurred by each of the multiple objective functions. A compelling example for considering the {\it min-max} approach is fair machine learning, where performance is typically evaluated across multiple demographic groups, which naturally induces a multi-objective optimization problem. While Pareto efficiency identifies trade-offs between groups, it may still permit highly unequal solutions. {\it min-max} objective effectively addresses this: it maximizes the minimum accuracy (or minimize the maximum error) across groups \cite{chen2020practical, chouldechova2020frontiers}).  See Appendix \ref{app:openQ} for more details on applications of  {\it min-max} OCO problem.

The formal  {\it min-max} OCO problem that we consider is as follows. Let there be 
two convex loss function sequences $f_t, g_t, t=1, \dots, T$ that are revealed after $\cA$ chooses action $x_t$ at time $t$. The {\it min-max}  cost of $\cA$ is 
\begin{equation}\label{defn:costAintro}
C_\cA = \max\left\{\sum_{t=1}^T f_t(x_t), \sum_{t=1}^T g_t(x_t) \right\}.
\end{equation}
 Two loss function sequences are considered  for simplicity of exposition since they already capture the basic difficulty. Extension to $K>2$ sequences is straightforward and detailed in Remark \ref{rem:multipleK}.
As a benchmark, we consider the static offline optimal algorithm $\opt$ that knows $f_t, g_t$ for $t=1, \dots, T$ at time $t=1$ itself but can only choose a single action $x^\star$ across all times
and the optimal cost is 
\begin{equation}\label{defn:optcostintro}
C_\opt = \max\left\{\sum_{t=1}^T f_t(x^\star), \sum_{t=1}^T g_t(x^\star) \right\},
\end{equation}
where $
x^\star = \arg \min_{x\in\cX} \max\left\{\sum_{t=1}^T f_t(x), \sum_{t=1}^T g_t(x) \right\}.$
Consequently, we define the {\it min-max} static regret of $\cA$ with actions $x_t$ as 
\vspace{-0.1in}
\begin{eqnarray} \label{intro-moregret-def}R_\cA(T) =\sup_{f_t, g_t} \{C_\cA - C_\opt\}.\end{eqnarray}
We refer to \eqref{intro-moregret-def} as the {\it min-max} OCO problem.
\vspace{-0.1in}
 \begin{remark}
The chosen benchmark in~\eqref{defn:optcostintro} is deliberately stringent and forces the algorithm to track both sequences globally and simultaneously
compared to $\min_{x \in \cX} \sum_{t=1}^T \max\{ f_t(x), g_t(x)\}$, for which the OGD algorithm applied on the surrogate function
    $h_t(x) = \max\{ f_t(x), g_t(x)\}$ would yield regret $O(\sqrt{T})$ independent of the number of distinct loss function sequences.
\end{remark}
\vspace{-0.34in}
\section{Related Work}
Early attempts at multi-objective OCO highlighted the difficulty of defining an appropriate performance metric \cite{mahdavi2013stochastic}. A common resolution, introduced in \cite{mahdavi2013stochastic}, is \emph{constrained online convex optimization} (COCO), where one objective is optimized while the remaining objectives are enforced as constraints.
In COCO, an algorithm chooses $x_t \in \mathcal{X}$ before observing a convex loss $f_t$ and convex constraint $g_t$, and performance is measured by both static regret with respect to a feasible comparator and cumulative constraint violation. COCO has been extensively studied under i.i.d. inputs \cite{mahdavi2013stochastic,uziel2017multi}, time-invariant constraints \cite{jenatton2016adaptive,mahdavi2012trading,yuan2018online,yi2021regret}, and fully time-varying constraints using primal--dual or drift-plus-penalty methods \cite{neely2017online,guo2022online,yu2017online}, with the best possible guarantee being regret of $O(\sqrt{T})$ and constraint violation of $O(\sqrt{T}\log T)$ with simple algorithms \cite{Sinha2024}, with further refinements yielding instance-dependent or improved violation bounds \cite{Vaze2025a,Vaze2025b}. Dynamic versions of COCO, where the benchmark varies over time, have also been considered \cite{chen2018bandit,cao2018online,vazecocowiopt2022,liu2022simultaneously}.

Despite superficial similarities, COCO and {\it min-max}  OCO are fundamentally different: COCO enforces feasibility over time, whereas {\it min-max} OCO compares against a static action that minimizes the worst total loss across objectives. Consequently, results for one setting do not directly imply results for the other.
A more detailed discussion on COCO can be found in Appendix \ref{app:COCO}.

\subsection{Global Cost Functions in \textsf{Experts} Setting}
A special case of the {\it min-max} OCO problem \eqref{intro-moregret-def} was considered in \cite{MannorGlobal}, where there are $K$ experts, with expert $i$'s loss at time $t$, being $\ell_t(i)$. If expert $i$ is selected by an algorithm $\cA$ with probability $p_t(i)$ at time $t$, then the overall expected loss for expert $i$ is $L_i = \sum_{t=1}^T p_t(i) \ell_t(i)$, and the cost of $\cA$ is the global cost   $C_{\cA}= \max_{i=1, \dots, K}\{ L_i\}$.
While \cite{MannorGlobal} achieved $O(\sqrt{T}\log K)$ regret using the {\textsf MULTI} algorithm, their setting fundamentally differs from ours: their objective is linear in the actions $p_t$, and the probability mass $p_t(i)$ is split among experts (summing to 1). In contrast, with \textit{min-max} OCO actions affect all objective functions simultaneously and in full. Consequently, results from \cite{MannorGlobal} and its extensions \cite{azar2014sequential, kesselheim2020online, rakhlinglobal} do not apply to our formulation, even for linear functions.
%

\subsection{Pareto Optimal Multi-Objective OCO} More recently, \cite{jiang2023multiobjective} attempted to study the multi-objective OCO from the Pareto optimality point of view. Towards this end, it used the concept of Pareto-suboptimality-gap (PSG), for $k=1, \dots, K$ loss functions $f_t^k$, and defined the regret to be\footnote{We write the equivalent simpler form given on Page 21  \cite{jiang2023multiobjective} compared to the original Definition 4 made on Page 5 of \cite{jiang2023multiobjective} that is mathematically more complicated.}
\vspace{-0.125in}
\begin{equation}\label{regretpareto}
\sup_{x^\star \in \cP^\star} \min_{k=1, \dots, K}\sum_{t=1}^T (f_t^k(x_t) -  f_t^k(x^\star)).
\end{equation}
where $\cP^\star$ is the set of points on the Pareto frontier for the $K$-dimensional vector 
$[\sum_{t=1}^T f_t^k]_{k=1}^K$. \cite{jiang2023multiobjective} proposed a mirror-descent inspired algorithm and 
showed that its regret \eqref{regretpareto} is $O(\sqrt{T})$ which notably is {\bf independent} of $K$.
The minimization over the $K$ possible loss functions in \eqref{regretpareto} made the regret \eqref{regretpareto} not depend on $K$, which highlights the weakness of the metric. Thus, even though \cite{jiang2023multiobjective} made a valid attempt at modelling the general multi-objective OCO, the weakness of the metric limited its
applicability.

\subsection{Per-slot {\it min-max} multi-objective optimization}
Compared to our `global' {\it min-max} objective \eqref{defn:optcostintro}, in \cite{lee2022online}, a {\bf per-slot} {\it min-max} multi-objective OCO was studied, where  at each round \(t = 1, \dots, T\): i) an online algorithm chooses \(x_t \in \mathcal{X}\), ii) the adversary chooses \(y_t \in \mathcal{Y}\), and iii) the online algorithm suffers a vector loss $
    \ell_t(x_t, y_t) = \big( \ell_{t,1}(x_t, y_t), \dots, \ell_{t,d}(x_t, y_t) \big) \in [-1,1]^d.$
The goal of an online algorithm is to solve $\min_{ \{x_t\} } \max_{j \in [d]} \sum_{t=1}^T \ell_{t,j}(x_t, y_t)$, knowing $\ell_t(.,.)$, but not $y_t$,
and its performance is compared against a per-round \emph{local minimax value} 
$w_L^t := \min_{x \in \mathcal{X}} \max_{y \in \mathcal{Y}} \max_{j \in [d]} \ell_{t,j}(x, y),$ 
where the cumulative benchmark is:
\begin{equation}\label{benchmarkdynamicminmax}
W_L^T := \sum_{t=1}^T w_L^t,\quad   \text{and regret} \quad
\mathrm{R}^{d}_\cA(T) := \max_{j \in [d]} \sum_{t=1}^T \ell_{t,j}(x_t, y_t) - W_L^T.
\end{equation}
A {\it Hedge}-type algorithm \cite{cesa2006prediction} was shown in \cite{lee2022online} to have $O(\sqrt{T})$ regret. 
Defining $\mathcal{Y} = \{e\}$ as a singleton, and $d=2$, such that $\ell_{t,1}(x_t, y_t)=f_t(x_t)$ and $\ell_{t,2}(x_t, y_t)=g_t(x_t)$ for $y_t=e$, we recover our model where two convex functions $f_t, g_t$ are revealed at time $t$, and for which  
$W_L^T = \sum_{t=1}^T  \min_{x_t\in \cX} \max\{f_t(x_t), g_t(x_t)\}).
$
This benchmark is fundamentally different than our considered benchmark \eqref{defn:optcostintro} since it is additive across time slots, and allows the benchmark to choose different actions across time slots. 
Even though this benchmark allows choosing different actions across time slots,  surprisingly, we show in Appendix \ref{app:benchmarkgap} that $W_L^T $  can be 
$3/2$ times more than our benchmark $C_\opt$ \eqref{defn:optcostintro} and hence algorithms with $O(\sqrt{T})$ regret \eqref{benchmarkdynamicminmax} do not 
yield a $O(\sqrt{T})$ {\it min-max} static regret \eqref{intro-moregret-def}.

\vspace{-0.175in}
\subsection{$\alpha$-Fairness}
Finally, we review the most closely related problem of \emph{online $\alpha$-fairness} studied by \cite{si2023enabling}. In this formulation with two users, and at each round $t\in[T]$ an online algorithm selects an action $x_t\in\mathcal X$. After $x_t$ is chosen, two nonnegative concave \emph{utility} functions $f_t,g_t:\mathcal X\to\mathbb{R}_{\ge 0}$ are revealed. Given a sequence of actions $\{x_t\}_{t=1}^T$, the time-averaged utilities are
$\bar f(x_{1:T}):=\frac{1}{T}\sum_{t=1}^T f_t(x_t)$, and $\bar g(x_{1:T}):=\frac{1}{T}\sum_{t=1}^T g_t(x_t)$. Let $\bar f(x_{1:T})=\bar f(x)$ when $x_t=x$ for all $t$.
Following \cite{si2023enabling}, performance is evaluated via the \emph{$\alpha$-fair utility}
\begin{equation}\label{defn:U}
U_\alpha(\bar f)+U_\alpha(\bar g),
\qquad
U_\alpha(z)=
\begin{cases}
\frac{z^{1-\alpha}}{1-\alpha}, & \alpha\neq 1,\\
\log z, & \alpha=1,
\end{cases}
\end{equation}
which is a concave aggregation of the two user utilities.
To compare with our {\it min-max} framework in this paper, define a cost function $\psi_\alpha:\mathbb{R}_{>0}^2\to\mathbb{R}$ where
$\psi_\alpha(a,b)
:=
-\,\Big(
U_\alpha(a)+U_\alpha(b)
\Big)$,
which is convex in $(a,b)$ for all $\alpha\ge 0$. The offline benchmark of \cite{si2023enabling} is then
$C_{\opt}(\alpha,T) = \min_{x\in\mathcal X}
\psi_\alpha( \bar f(x), \bar g(x))$ and the cost of $\cA$ is $C_{\cA}(\alpha,T) = \psi_\alpha(\bar f(x_{1:T}), \bar g(x_{1:T}))$.

In the limit $\alpha\to\infty$  costs $C_{\cA}(\alpha,T)$ and $C_{\opt}(\alpha,T)$ become equivalent to the {\it min-max} costs defined in \eqref{defn:costAintro} and \eqref{defn:optcostintro}, respectively. Thus, the \emph{online $\alpha$-fairness} problem~\cite{si2023enabling} formally subsumes our {\it min-max} OCO problem as the limiting case $\alpha\to\infty$.

~\cite{si2023enabling}  show that with adversarial input, the regret $
C_{\cA}(\alpha,T)-C_{\opt}(\alpha,T) =\Omega(T)$ for any online algorithm, even with fixed $\alpha$, ruling out sublinear regret in this model. Motivated by this impossibility, they consider a stochastic (smoothed) input model, under which they propose an online horizon-fair (OHF) policy and prove that its regret is $O(\alpha\sqrt{T})$. Notably, since the regime of interest for our {\it min-max} objective~\eqref{defn:costAintro} corresponds to $\alpha\to\infty$, the stochastic regret guarantee of~\cite{si2023enabling} {\bf becomes vacuous}. See Remark \ref{rem:sisalem} for more details.




\subsection{Our Contributions}
\begin{enumerate}[leftmargin=*]

\item We begin by deriving a lower bound of $\Omega(T)$ on the {\it min-max} static regret \eqref{intro-moregret-def} of any online algorithm with 
adversarial input. The input construction is inspired from  \cite{si2023enabling} and is modified to suit our specific problem.

 \item 
Motivated by the impossibility of sublinear {\it min-max} static regret in the adversarial setting, we consider a stochastic \textbf{i.i.d.} model,\footnote{We also consider the Martingale difference and Markov input models in Appendix \ref{app:martingale} and \ref{app:markov}.} where loss pairs $(f_t, g_t)$ are drawn from a fixed unknown distribution $\cD$. This setting is standard for applications like fair resource allocation and machine learning and remains analytically challenging.

Under the \textbf{i.i.d.} model, the benchmark is $\bbE\{C_\opt\}$ where $C_\opt$ has been defined in \eqref{defn:optcostintro}. Directly analyzing the
{\it min-max} static regret \eqref{intro-moregret-def} with respect to $\bbE\{C_\opt\}$ is technically
challenging, since $C_\opt$ depends on the entire realized sequence of losses.
Instead, we introduce a deterministic surrogate benchmark $\bar C_\opt$ obtained by averaging out the randomness
in the losses:
\begin{equation}\label{intro:eq:optminmax}
\bar C_\opt
= T \cdot \min_{x\in\mathcal X}\max_{\theta\in\Delta_2}\theta^\top \mu(x),
\qquad
\mu(x) = \bbE\{[f_t(x), g_t(x)]\},
\end{equation} 
where $\Delta_K$ is the $K$-dimensional simplex. Importantly, as shown in Appendix~\ref{app:regretconnection}, $$\bbE\!\left\{\,\big|C_\opt - \bar C_\opt\big|\,\right\}
= O\!\left(\sqrt{T \log T}\right).\footnote{With $K$ sequences, the bound becomes
$O\!\left(\sqrt{T \log (TK)}\right)$.}$$

Using $\bar C_\opt$ as the benchmark, we define the \emph{expected {\it min-max} static regret} of an
online algorithm $\cA$ as
\begin{equation}\label{defn:expregretintro}
\bar R_{\cA}(T)
= \sup_{\cD} \left\{ \bbE\{C_{\cA}\} - \bar C_\opt \right\}.
\end{equation}
For clarity, we refer to regret measured with respect to $\bbE\{C_\opt\}$ as the
\emph{true {\it min-max} static regret}, and regret measured with respect to $\bar C_\opt$ as the
\emph{expected {\it min-max} static regret}.

In the remainder of the paper, we bound $\bar R_{\cA}(T)$. From the above concentration bound, the true {\it min-max} static regret satisfies
$
\bbE\{C_{\cA}\} - \bbE\{C_\opt\}
\;\le\;
\bar R_{\cA}(T) + O\!\left(\sqrt{T \log T}\right)$.
\footnote{With $K$ sequences, the additive term becomes
$O\!\left(\sqrt{T \log (TK)}\right)$.}

Even under the \textbf{i.i.d.} model, the lower bound on the  true {\it min--max} static regret is 
$\Omega(\sqrt{T})$. This follows by taking $f_t \equiv g_t$ for all $t$, which reduces the
problem to standard OCO, for which
$\bbE\{R^{\text{single}}_{\cA}(T)\} = \Omega(\sqrt{T})$ even under i.i.d.\ inputs (e.g., random
linear losses; see \cite{HazanBook}).

\item \textbf{Algorithm}
While considering expected {\it min-max} static regret \eqref{defn:expregretintro}, the structure of \eqref{intro:eq:optminmax} reveals the core intuition for our approach: the optimal policy corresponds to a saddle point $(x^\star, \theta^\star)$ that solves ${\bar C}_\opt$ in \eqref{intro:eq:optminmax}. Since the distribution $\cD$ is unknown, we cannot compute this pair offline. Instead, we propose a natural online algorithm (Algorithm \ref{alg:main}) that interlaces two classic strategies: 
i) {\it Hedge} to update the weights $\theta_t$ (solving the inner maximization over objectives in \eqref{intro:eq:optminmax}), and 
ii) OGD to update the action $x_t$ (solving the outer minimization on the surrogate loss $\theta_t^\top [f_t, g_t]$ in \eqref{intro:eq:optminmax}). 

While \eqref{intro:eq:optminmax} resembles a zero-sum game \cite{pmlr-v97-cardoso19a}, the problem is fundamentally different: here, the learner controls \emph{both} $x_t$ and $\theta_t$ to track an unknown distribution. The novelty of our approach lies not in the individual choice of {\it Hedge} or OGD, but in their principled combination which simplifies an otherwise intractable analysis. While other candidates like FTRL or algorithms for COCO \cite{yu2017online, Sinha2024} have an intuitive appeal for minimizing \eqref{defn:expregretintro}, there are obvious analytical roadblocks caused by the global $\max$ in the cost function \eqref{defn:costAintro} even with stochastic i.i.d. input. Our framework resolves this difficulty through a modular regret decomposition and decouples the dynamics of {\it Hedge} and OGD that renders the proof both modular and elegant. 

We prove that our algorithm achieves an expected {\it min-max} static regret \eqref{defn:expregretintro} of  $O(\sqrt{T})$ (or $O(\sqrt{T \log (K)})$ for $K$ objectives). Analogous results for Martingale difference and Markov input models are provided  in Appendix \ref{app:martingale} and \ref{app:markov}.

Recall that the $\alpha$-fair model of \cite{si2023enabling} is equivalent to our {\it min-max} setting when $\alpha\rightarrow \infty$. Thus, importantly, 
compared to the $O(\alpha\sqrt{T})$ regret bound of \cite{si2023enabling}
we obtain an $\alpha$-independent regret bound.

\item {\bf With Bandit Feedback:} Exploiting our modular proof approach, we easily extend our results to the bandit setting, where only function evaluations at the most recent actions are available  without any gradient feedback. 
We show that the expected {\it min-max} static regret \eqref{defn:expregretintro} is $O(T^{3/4}\sqrt{ \log K})$ and $O(\sqrt{T \log K})$ for one-point and two-point bandit feedback, respectively, matching the best known bounds on OCO in the respective bandit settings.
\item 

We also consider the {\it min-max} generalization of the {\it universal prediction of binary sequences} \cite{feder2002universal, cover1966behavior}, where at each
time $t$,  an online algorithm $\cA$ has to choose a binary action $x_t$ before two binary sequences $b_t^1, b_t^2$'s are revealed. The cost of $\cA$ on  sequence $k$ is its Hamming distance between $\{x_t\}_{t=1}^T$ and $\{b_t^k\}_{t=1}^T, k=1,2$, and the overall cost of $\cA$ is the maximum of its cost on the two sequences. 
Even for this case, we show that the {\it min-max} static regret of any online algorithm is $\Omega(T)$ with 
adversarial input. With stochastic input where $(b_t^1, b_t^2)$ are i.i.d. across time, we propose an algorithm with $O(1)$ true {\it min-max} static regret. In comparison, for the classical version \cite{feder2002universal, cover1966behavior} with only a single sequence $b_t^1$, Follow-the-Leader (FTL) algorithm achieves $O(1)$ regret when 
the input is i.i.d. \cite{feder2002universal}, while  \cite{cover1966behavior} achieves  $O(\sqrt{T})$ regret in the adversarial case.

\end{enumerate}

We discuss the applications of multi-objective OCO and research questions left open in Appendix \ref{app:openQ}, while numerical results are provided in Appendix \ref{sec:sim}.

\vspace{-0.2in}
\section{Assumptions}\label{sec:sysmodel}
We make the following  assumptions that are standard in  the OCO  literature  \cite{HazanBook}.

\begin{assumption}[Convexity] \label{cvx}
$\mathcal{X} \subset \bbR^d$ is the admissible set that is closed, convex and has a finite Euclidean diameter $D$.  
Cost functions $f_t, g_t: \mathcal{X} \mapsto \mathbb{R}$  are convex for all $t\geq 1$, and are bounded $
|f_t(x)|\le B, |g_t(x)|\le B, \ \forall \ x \in \cX$.  
\end{assumption}
\begin{assumption}[Lipschitzness] \label{bddness}
 All cost functions $\{f_t,g_t\}_{t\geq 1}$  are $G$-Lipschitz, i.e., for any $x, y \in \mathcal{X},$ we have 
$ 	|f_t(x)-f_t(y)| \leq G||x-y||,~
 	|g_{t}(x)-g_{t}(y)| \leq G||x-y||, ~\forall t\geq 1$.
	\end{assumption}

\begin{assumption} {\bf Information Structure:}\label{defn:information} On round $t,$ the online algorithm $\cA$ first chooses an admissible action $x_t \in \mathcal{X}$ 
 and then the adversary chooses two convex cost functions $f_t: \mathcal{X} \to \mathbb{R}$ and $g_t: \mathcal{X} \to \mathbb{R}$. 
 Once the action $x_t$ has been chosen, both $f_t(x_t), \nabla f_t(x_t)$ and  $g_t(x_t), \nabla g_t(x_t)$ are revealed, as is standard in the literature. We will also consider the bandit information structure in Section \ref{sec:bandit} where no gradient information is available.
 \end{assumption}



\section{Warmup}\label{sec:warmup}
In this section, we highlight the difficulty in  solving the {\it min-max} OCO Problem \eqref{intro-moregret-def} in the adversarial input setting. 
To illustrate the basic difficulty, 
we even allow an online algorithm to know $f_t, g_t$ completely {\bf before} 
choosing its action $x_t$ at time $t$. Under this enhanced information setting, consider the greedy (locally optimal) algorithm that chooses 
$x_t = \min_{x\in \cX}\max\{f_t(x), g_t(x)\}.$ Recall that when $g_t=f_t \ \forall \ t$, i.e. in the OCO setting, this algorithm will have non-positive regret $R^{\text single}_\cA(T)$  since $f_t(x_t) \le f_t(x)$ for any $x$. However, as we show next, the greedy algorithm suffers 
a linear {\it min-max} static regret \eqref{intro-moregret-def}.
Let $T=2N$ and  for $j=1, \dots, N$, consider the following linear functions on $\cX =[0,1]$,
\vspace{-0.1in}
\begin{align}\nn
f_{2j-1}(x) =1.2-0.2x, & \qquad f_{2j}(x)  =x,\\ \label{eq:lb1}
\hspace{-1in}g_{2j-1}(x)  =x, &\qquad  g_{2j}(x)  =0.8+0.2x.
\end{align}
Clearly, $
\arg\min_{x\in \cX}\{\max(f_{2j-1}(x),g_{2j-1}(x))\}  =1 \  \text{and} 
\arg\min_{x\in \cX}\{\max(f_{2j}(x),g_{2j}(x))\}  =0.$

Thus, the greedy algorithm chooses $x_{2j-1}=1$ and $x_{2j}=0,$ and the cost $C_\cA$ \eqref{defn:costAintro} of the greedy algorithm is $1.8N$.
However, if an algorithm $\cA_1$ chooses a single static action $x^\star=0$ for all times, its cost $C_{\cA_1}$ \eqref{intro-moregret-def} is $1.2N$, and hence
$C_\opt \le 1.2N.$
Thus, the {\it min-max} static regret \eqref{intro-moregret-def} of the greedy algorithm is linear in $T$. 
\section{Lower bound on {\it min-max} static regret \eqref{intro-moregret-def} for all online algorithms}\label{sec:lb}
We formalize the intuition presented in Section \ref{sec:warmup} as follows.
\begin{theorem}\label{thm:lb} The  {\it min-max} static regret \eqref{intro-moregret-def} for any online algorithm is $\Omega(T)$.
\end{theorem}
The proof of Theorem \ref{thm:lb} can be found in Appendix \ref{app:lb}. The lower bound derived in Theorem \ref{thm:lb} uses the fact that  both $f_t$ and $g_t$ are convex and not strongly convex. 
With strongly convex functions, the lower bound may weaken.
In light of Theorem \ref{thm:lb}, next, we consider the stochastic i.i.d. case for solving the {\it min-max} OCO problem which is also non-trivial.

\section{Stochastic i.i.d. input model}\label{sec:iid}
We consider the i.i.d. input model, where 
at time $t$, loss function pair $(f_t, g_t)$ is chosen independently and is identically (jointly) distributed as $\cD$ (unknown) across all $t$. 
 The objective is to minimize the expected  {\it min-max} static regret \eqref{defn:expregretintro}.

\begin{remark}
We note that only two loss function sequences are considered for simplicity of exposition since it captures the basic difficulty of the problem, and all the results that we derive apply directly to the case when there are $K$ function sequences (see details in Remark \ref{rem:multipleK}).
\end{remark}

\subsection{Algorithm}
In this section, we present an algorithm to solve the i.i.d.  {\it min-max} OCO problem \eqref{defn:expregretintro}.

\begin{algorithm}[H]
\caption{{\it Hedge}+OGD}
\label{alg:main}
\begin{algorithmic}[1] 
\State \textbf{Input:} step sizes $\eta_{x,t} > 0$, $\eta_{\theta,t} > 0$, feasible set $\mathcal{X}$
\State \textbf{Initialize:} $w_1 = (1,1)$, $\theta_1 = w_1 / \|w_1\|_1$, $f_0\equiv 0, g_0\equiv 0$ and play arbitrary $x_0 \in \mathcal{X}$

\For{$t=1,\dots,T$}
  \State Observe $(f_{t-1}(x_{t-1}),g_{t-1}(x_{t-1}))$ and $(\nabla f_{t-1}(x_{t-1}), \nabla g_{t-1}(x_{t-1}))$
  \State Form convex function $x \mapsto {\tilde \Lambda}_t(x,\theta_t) := \theta_{t,1} f_{t-1}(x) + \theta_{t,2} g_{t-1}(x)$
  \State {\bf OGD:} $y_t = x_{t-1} - \eta_{x,t} \nabla {\tilde \Lambda}_t(x_{t-1},\theta_t)$ and find $x_t = \text{Proj}_{\mathcal{X}}(y_t)$
  \State {\bf Play action $x_t$} \quad Evaluate $\lambda_t = (f_t(x_t), g_t(x_t))$
  \State {\it Hedge:} $w_{t+1,i} = w_{t,i} \exp(\eta_{\theta,t} \lambda_{t,i}), i=1,2$ 
  \State Normalize $\theta_{t+1} = w_{t+1}/\|w_{t+1}\|_1$
\EndFor
\end{algorithmic}
\end{algorithm}

Towards that end, if suppose an online algorithm knew the optimizing $\theta^\star$ in \eqref{intro:eq:optminmax}, then by executing an OGD algorithm for OCO with a single function $\theta_1^\star f_t + \theta_2^\star g_t$ at time $t+1$, it would achieve a regret of $O(\sqrt{T})$ using standard results \cite{HazanBook}. However,  the optimizing $\theta^\star$ that depends on $\cD$ (unknown) remains unknown, but, this 
observation motivates our {\it algorithm} that constructs \newline i) $\theta_t$ at time $t$ as a surrogate for $\theta^\star$ using the {\it Hedge} algorithm (gains version) treating the two function sequences $\{f_t\}_{t=1}^T$ and $\{g_t\}_{t=1}^T$ as the gains of the two experts (see Section \ref{sec:hedgeintro} for basic definition of the Experts Problem and the {\it Hedge} algorithm), and \newline ii) for each fixed value of 
$\theta_t$, {\bf plays the action} chosen by OGD applied to the single surrogate function $\theta_{t1} f_t + \theta_{t2} g_t$ at time $t+1$. Pseudocode is given in Algorithm \ref{alg:main}.

\subsection{Guarantee}

\begin{theorem}\label{thm:main} Under Assumptions \ref{cvx} and \ref{bddness}, for $\eta_{\theta,t}\le  \sqrt{\frac{2\ln 2}{B^2t}}, \eta_{x,t} = \frac{D}{G\sqrt{t}}
$, the expected {\it min-max} static regret \eqref{defn:expregretintro} of the proposed algorithm (Algorithm \ref{alg:main}) is $${\bar R}_\cA(T) = O(\sqrt{T}).$$
\end{theorem}

\begin{proof}
The main ingredient of the proof as well as the inspiration for Algorithm \ref{alg:main} is the following decomposition \eqref{defn:regretdecomp} of the expected {\it min-max} static regret \eqref{defn:expregretintro}. Let $\Delta_K$ be the $K$-dimensional simplex. Recall the definition of 
$\Lambda_t(x,\theta):=\theta_1 f_t(x)+\theta_2 g_t(x)$, where $\theta = (\theta_1, \theta_2) \in \Delta_2$.
By adding and subtracting two terms, $\sum_{t=1}^T \Lambda_t(x_t,\theta_t)$ and $\min_{x\in\mathcal X}\sum_{t=1}^T \Lambda_t(x,\theta_t)$ inside the expectation   in the RHS of \eqref{defn:expregretintro}, we write the expected {\it min-max} static regret \eqref{defn:expregretintro} as 
\vspace{-0.1in}
\begin{equation}\label{defn:regretdecomp}
{\bar R}_\cA(T) = \sup_{\cD}\{\bbE\{R_1\} + \bbE\{R_2\} + \bbE\{R_3\}\} \le \sup_{\cD}\{\bbE\{R_1\}\}  + \sup_{\cD}\{\bbE\{R_2\}\} + \sup_{\cD}\{\bbE\{R_3\}\},
\end{equation}
where 
\[
R_1=\max_{\theta\in\Delta_2}\sum_{t=1}^T \Lambda_t(x_t,\theta)-\sum_{t=1}^T \Lambda_t(x_t,\theta_t),\]
\[  R_2=\sum_{t=1}^T \Lambda_t(x_t,\theta_t)-\min_{x\in\mathcal X}\sum_{t=1}^T \Lambda_t(x,\theta_t),
\]
and
\vspace{-0.1in} \[
R_3 =\min_{x\in\mathcal X}\sum_{t=1}^T \Lambda_t(x,\theta_t)-{\bar C}_\opt = \min_{x\in\mathcal X}\sum_{t=1}^T \Lambda_t(x,\theta_t)- T\cdot \min_{x\in\mathcal X}\max_{\theta\in\Delta_2}\theta^\top\mu(x) .
\]
The interpretation of $R_1$ is that it is the regret for the gains version of $2$-Experts Problem \cite{cesa2006prediction}, where the two function sequences $f_t,g_t$ are the gains of two experts, with respect to an optimal $\theta^\star \in \Delta_2$ for fixed actions $x_t$ across time slots. This motivates the use of {\it Hedge} algorithm  (gains version) in our proposed algorithm that is tailor-made for updating $\theta_t$ for fixed actions $x_t$. 

Similarly, $R_2$ is the static regret for the OCO problem relative to static optimal action $x^\star$ for a single surrogate objective function sequence of $\Lambda_t(x,\theta_t) = \theta_{t1} f_t(x)+\theta_{t2} g_t(x)$ for a fixed $\theta_{t}$. Hence using an OGD algorithm is natural for this purpose as done in the algorithm for each time $t$, with a fixed weight $\theta_t$.

The third term $R_3$ is more nuanced and captures the difference between the cost of $\opt$ for a fixed sequence of $\theta_t, t=1,\dots, T$ and the optimal benchmark cost ${\bar C}_\opt$ \eqref{intro:eq:optminmax}, respectively.

We  bound $R_1$ and $R_2$ sample-path wise for each realization of $f_t, g_t$, while for $R_3$ we will bound $\bbE\{R_3\}$ as follows whose proofs are provided in Appendix \ref{proofLemmaR1}, \ref{proofLemmaR2}, and \ref{proofLemmaR3}, respectively
\begin{lemma}\label{lem:R1} For $\eta_{\theta,t}\le \sqrt{2\ln 2/(B^2t)}$, $\sup_{\cD}\{R_1\} \le O(\sqrt{T})$. 
\end{lemma}

\begin{lemma}\label{lem:R2} Choosing $\eta_{x,t} = \frac{D}{G\sqrt{t}}$, $\sup_{\cD}\{R_2\} \le O(\sqrt{T})$. 
\end{lemma}

\begin{lemma}\label{lem:R3} $\bbE\{R_3\} \le 0$. 
\end{lemma}


Combining Lemma \ref{lem:R1}, \ref{lem:R2} and \ref{lem:R3}, and the regret decomposition \eqref{defn:regretdecomp}, 
we get that the expected {\it min-max} static regret ${\bar R}_\cA(T) = O(\sqrt{T})$ \eqref{defn:expregretintro}  with
$\eta_{\theta,t}\le  \sqrt{\frac{2\ln 2}{B^2t}},  \text{and}\  \eta_{x,t} = \frac{D}{G\sqrt{t}}$.
\end{proof}
\vspace{-0.3in}
\begin{remark}
The regret decomposition \eqref{defn:regretdecomp} does multiple things at the same time i) identifies the critical difficulty of the problem, ii) motivates a natural algorithm, and ii) keeps the 
proof modular, where the  regret bound for each module is simple and elegant, almost first principles based.  
\end{remark}


\begin{remark}
Although the proof of Theorem \ref{thm:main} is remarkably simple and elegant, it hinges on the critical regret decomposition \eqref{defn:regretdecomp} -- which is 
not obvious a priori. This decomposition eliminates the need for sophisticated technical machinery needed in related prior work \cite{mahdavi2013stochastic} that also considers the i.i.d. input and highlights the underlying simplicity of the result. 
We would like to emphasize that the elegance of the argument should not obscure its importance.
\end{remark}

\begin{remark} For the $\alpha$-fair problem, \cite{si2023enabling} derived a regret guarantee of $O(\alpha \sqrt{T})$, which becomes vacuous for our regime of interest since $\alpha$-fair problem is equivalent to our {\it min-max} setting when $\alpha \rightarrow \infty$. Compared to \cite{si2023enabling}, Algorithm \ref{alg:main}, 
has an $\alpha$-independent regret guarantee. It is  tempting to use Algorithm \ref{alg:main} for the $\alpha$-fair problem for finite  $\alpha$, however, as we point out in Remark \ref{rem:sisalem}, the $\alpha$-fair problem with finite $\alpha$ and $\alpha \rightarrow \infty$ are fundamentally different. 
\end{remark}

\begin{remark}\label{rem:multipleK} When there are $K$ loss function sequences $f_t^k, k=1, \dots, K$ that are chosen i.i.d. from $\cD$,  define $\theta \in \Delta_K$, (where $\Delta_K$ is the $K$-dimensional simplex) and  
$\Lambda_t(x,\theta) =\sum_{k=1}^K \theta_k f_t^k$.
With $\mu(x) = \bbE\{f_t^1(x), \dots, f_t^K(x)\}$, analogous to \eqref{defn:expregretintro}, the expected
{\it min-max} static regret of $\cA$ is \begin{equation} \label{intro-moregret-def-multiple}
{\bar R}_\cA(T) =\sup_{\cD}\bbE\left\{\max_{\theta \in \Delta_K} \sum_{t=1}^T \Lambda_t(x_t,\theta) - T\cdot \min_{x\in\mathcal X}\max_{\theta\in\Delta_K}\theta^\top\mu(x)\right\}.
\end{equation}

With this new definition of $\theta$ and $\Lambda_t$, Algorithm \ref{alg:main} remains as is with {\it Hedge} choosing $\theta_t \in \Delta_K$ while 
$x_t$ the single action chosen by OGD with respect to the surrogate loss function $\Lambda_t(x,\theta_t)$ with fixed $\theta_t$. Assuming all $f_t^k, k=1, \dots, K$ satisfy Assumption \ref{cvx} and \ref{bddness},  the only change in the guarantee with $K>2$ compared to Theorem \ref{thm:main} with $K=2$ is because of Lemma \ref{lem:R1} that is now $O(\sqrt{T \log K})$ \cite{HazanBook} since there are $K$ experts corresponding to the $K$ loss function sequences $f_t^k, k=1, \dots, K$ . The OGD guarantee of Lemma \ref{lem:R2} stays the same since $\|\nabla \Lambda_t(x_t,\theta)\| \le G$ as before. Moreover, Lemma \ref{lem:R3} holds as is. Thus, the overall expected {\it min-max} static regret of Algorithm \ref{alg:main} with $K$ loss function sequences is 
$O(\sqrt{T \log K}).$
\end{remark}


\begin{remark}\label{rem:strongcvx}[Extension to Strongly Convex $f_t$'s and $g_t$'s]
One limitation of our proposed algorithm and the proof is that because of the use of the {\it Hedge} algorithm for updating $\theta_t$, the regret $\sup_{\cD}\{R_1\}$ can at best be  $O(\sqrt{T})$ with stochastic i.i.d. input without any further assumptions. Thus, unlike OCO, where OGD algorithm achieves logarithmic regret in $T$ when $f_t$'s are strongly convex, our algorithm's regret is only $O(\sqrt{T})$ even when both $f_t, g_t$ are strongly convex. Is this a limitation of the algorithm or of that of the problem itself remains an interesting unresolved question.
\end{remark}


\section{Extension to Bandit Feedback}\label{sec:bandit}
We now analyze the expected {\it min-max} OCO problem~\eqref{defn:expregretintro} under bandit feedback: at each time $t$, after choosing $x_t$, the algorithm observes only the scalar losses $\{f_{t-1}^k(\cdot)\}_{k=1}^K$ at one point (one-point model) or two points (two-point model), and receives no gradient information. 
For this bandit model, we consider a simple adaptation of Algorithm \ref{alg:main} by replacing its  OGD part with the bandit information counterpart, while keeping the {\it Hedge} part as it is, given by Algorithm \ref{alg:bandit}. 
Exploiting the modular nature of the proof in Section \ref{sec:iid}, the results follow almost immediately.

\begin{algorithm}[H]
\caption{Bandit-{\it min-max}}
\label{alg:bandit}
\begin{algorithmic}[1]
\State \textbf{Input:}{Feasible set $\mathcal X\subset\mathbb R^d$, parameter $\{\delta, \rho\}$,
stepsizes $\{\eta_{x,t}\},\{\eta_{\theta,t}\}$,
$\mathsf{mode}\in\{\text{one-point},\text{two-point}\}$}
Initialize $w_1=(1, 1, \dots, 1)_K$, $\theta_1=w_1/\|w_1\|_1$, choose $x_1 \in (1-\rho)\mathcal X$\;

\For{$t=1,\dots,T$}
  Sample $u_t\sim\mathrm{Unif}(\mathbb S^{d-1})$\;
  \If{$\mathsf{mode}=\text{one-point}$}
    Query $\tilde x_{t} = x_{t} + \delta u_t$\;
    Observe $\{f_{t}^k(\tilde x_{t})\}_{k=1}^K$\;
    $\widehat\nabla_t = \frac{d}{\delta} \Lambda_{t}(\tilde x_{t},\theta_{t})u_t$ where 
    $\Lambda_{t}(\tilde x_{t},\theta_{t}) = \sum_{k=1}^K \theta_{t,k} f_{t}^k(\tilde x_{t})$\;
    $\ell_{t,k} = f_{t}^k(\tilde x_{t})$\;
  
  \Else
    Query $x_{t}^{+} = x_{t} + \delta u_t$, $x_{t}^{-} = x_{t} - \delta u_t$\;
    Observe $\{f_{t}^k(x_{t}^{+})\}_{k=1}^K$ and $\{f_{t}^k(x_{t}^{-})\}_{k=1}^K$\;
    
    $\widehat\nabla_t = \frac{d}{2\delta} \big(\Lambda_{t}(x_{t}^{+},\theta_{t}) - \Lambda_{t}(x_{t}^{-},\theta_{t})\big)u_t$\;
    $\ell_{t,k} = \tfrac12 \left( f_{t}^k(x_{t}^{+}) + f_{t}^k(x_{t}^{-}) \right)$\;
  \EndIf

  $x_{t+1} = \text{Proj}(x_{t} - \eta_{x,t}\widehat\nabla_t, (1-\rho)\mathcal{X})$\;
  $w_{t+1} = w_{t}\exp(\eta_{\theta,t}\ell_{t})$\;
  $\theta_{t+1} = w_{t+1}/\|w_{t+1}\|_1$\;
\EndFor
\end{algorithmic}
\end{algorithm}

\begin{theorem} Let $\sfr$ be such that $r\cB_1\subseteq \cX \subseteq D\cB_1$ and $\cB_1$ is the unit ball.
\label{thm:bandit}
\begin{enumerate}[leftmargin=0pt]
    \item (\textbf{One-point feedback}) 
    With $\delta = T^{-1/4}$, $\eta_{x,t}= D/(G d^{1/2} t^{3/4})$, $\rho=\delta/\sfr$,
    and $\eta_{\theta,t}= \sqrt{\log K}/(B\sqrt{t})$, expected {\it min-max} static regret 
      $ {\bar R}_\cA(T) \leq c_1 \, d^{1/2} G D \, T^{3/4} \;+\; c_2 B \sqrt{T \log K}$,
    for constants $c_1,c_2>0$.
    \item (\textbf{Two-point feedback}) 
   With $\delta = T^{-1/2}$, $\rho=\delta/\sfr$,
    $\eta_{x,t}= D/(G d^{1/2}\sqrt{t})$, and 
    $\eta_{\theta,t}= \sqrt{\log K}/(B\sqrt{t})$,  expected {\it min-max} static regret  satisfies
       $ {\bar R}_\cA(T) \leq c_3 \, d^{1/2} G D \, \sqrt{T} \;+\; c_4 B \sqrt{T \log K}$,
    for  constants $c_3,c_4>0$.
\end{enumerate}
\end{theorem}
Proof of Theorem \ref{thm:bandit} can be found in Appendix \ref{app:banditproof}.

{\it Discussion:} Compared to the limited information setting considered in Section \ref{sec:sysmodel}, where gradient information is available, with bandit feedback 
the expected {\it min-max} regret guarantees suffer, however, the degraded guarantees match the best known guarantees for  OGD in the bandit setting. The proof of the result is simple because of  the modularity of our algorithm and the regret decomposition that decomposes the total regret into regret resulting from {\it Hedge} and OGD, and the fact that only regret for OGD changes with bandit feedback.

\vspace{-0.1in}
\section{Binary Prediction Problem}
\label{sec:binary}
In this section, we consider the {\it min-max} {\it universal prediction of binary sequences}, where at each
time $t$, two binary sequences
$b_t^1,b_t^2\in\{0,1\}$ are revealed after an online algorithm $\cA$ (predicts) chooses an action $x_t\in\{0,1\}$, and cost of $\cA$ is
\vspace{-0.1in}
$$C^b_\cA\;=\;\max\left\{\sum_{t=1}^T\mathbf 1\{x_t\neq b_t^1\},\; \sum_{t=1}^T\mathbf 1\{x_t\neq b_t^2\}\right\}.$$
Similar to \eqref{defn:optcostintro}, 
\vspace{-0.1in}
$$C^b_\opt
\;=\;
\min_{x\in\{0,1\}}
\max\Big\{
\sum_{t=1}^T \mathbf 1\{x\neq b_t^1\},\;
\sum_{t=1}^T \mathbf 1\{x\neq b_t^2\}
\Big\},$$
and the regret is defined as $R^b_\cA(T)  = \max_{b_t^1,b_t^2}\{C^b_\cA-C^b_\opt\}.$

 Similar  to Theorem \ref{thm:lb}, we show that the {\it min-max} static regret $R^b_\cA(T)$ for any online algorithm is $\Omega(T)$ with adversarial input in contrast to the classical setting \cite{cover1966behavior, feder2002universal} where with a single binary sequence the regret is $O(\sqrt{T})$ with adversarial input.
\begin{theorem}\label{thm:lbbinary} The  {\it min-max} static regret $R^b_\cA(T)$ for any online algorithm is $\Omega(T)$.
\end{theorem}
The proof of Theorem \ref{thm:lbbinary} can be found in Appendix \ref{app:lbbinary}. 
Thus, once again we consider the stochastic i.i.d. input as follows.

\subsection{$(b_t^1, b_t^2)$ i.i.d. across time}
In this section, we consider that $(b_t^1,b_t^2)$ are generated from an unknown distribution with i.i.d. across time. Under this setting, we can directly use Theorem \ref{thm:main} to show that Algorithm \ref{alg:main} can be used to get a true {\it min-max} static regret bound of $\bbE\{R^b_\cA(T)\} =O(\sqrt{T\log T})$, see details in Appendix \ref{app:binaryspecialcase}.

Next, we show that in this binary case,  we can do better than Algorithm \ref{alg:main} and get true {\it min-max} static regret bound $\bbE\{R^b_\cA(T)\} = O(1)$ without involving the averaged-out benchmark ${\bar C}_\opt^b$ that is defined similar to  \eqref{intro:eq:optminmax}.

The main challenge in this binary {\it min-max} prediction setting arises in
\emph{disagreement regimes}, where the two sequences $\{b_t^1\}_{t=1}^T, \{b_t^2\}_{t=1}^T$ favor opposite actions.
In such cases, simply following a majority or resolving ties arbitrarily can
lead to persistent oscillations, since each sequence individually advocates a
different decision. To achieve low expected {\it min-max} static regret, an algorithm must determine
which of these conflicting preferences induces the larger worst-case cost and
commit to that action.

To address this we propose \textbf{Adaptive Stronger-Bias Double Majority (ASB-DM)} algorithm (Algorithm~\ref{alg:binary} in Appendix~\ref{app:dm}) that exploits the \emph{strength} of the statistical evidence,
rather than only its direction. When the empirical preferences of the two
sequences disagree, ASB-DM aggregates the evidence by comparing the combined
bias toward the two actions and selects the one with the smaller inferred
worst-case loss. As the empirical estimates concentrate, this stronger-bias
rule allows the algorithm to reliably distinguish the optimal action from the
suboptimal one, ensuring that the decision stabilizes and the total number of
mistakes remains finite.

\begin{theorem}\label{thm:binary}
Assume the pairs $(b_t^1, b_t^2)$ are generated i.i.d. across time $t$ from a joint distribution with marginals $b_t^1 \sim \mathrm{Ber}(q_1)$ and $b_t^2 \sim \mathrm{Ber}(q_2)$. If the preferences are asymmetric (i.e., $q_1, q_2 \neq 1/2$ and $q_1 + q_2 \neq 1$), then the true {\it min-max} regret of Algorithm \ref{alg:binary} satisfies $\bbE\{R^b_\cA(T)\}  = O(1)$.
\end{theorem}
%
Proof of Theorem \ref{thm:binary} is provided in Appendix \ref{app:dm}. Note that extension to the case when there are $K\ge 2$ sequences $b_t^k, k=1,\dots, K$ is immediate with $O(1)$ term scaling linearly with $K$.

\begin{remark}[Singularity at $q_1+q_2=1$]
The necessity of the condition $q_1+q_2 \neq 1$ is best illustrated by the example where $q_1 = 0.4$ and $q_2 = 0.6$. To determine the static optimal action, we compare the worst-case expected costs: action $0$ incurs $0.6T$ (driven by Sequence 2), which is identical to the $0.6T$ cost incurred by action $1$ (driven by Sequence 1).
Since the worst-case costs are identical, the static optimal action is not unique. 
At this singularity, any online algorithm faces an inherent lack of statistical separation between the actions. Because the expected costs are perfectly balanced, empirical observations will fluctuate around the true parameters due to random sampling noise. Without a strictly positive gap to stabilize the decision process, any algorithm will be forced to switch between actions repeatedly—effectively behaving as a symmetric random walk—leading to an accumulation of regret at a rate of $\Omega(\sqrt{T})$. 
\end{remark}

%
%

%

\section{Conclusions}
In this paper, we have  made a basic advance in the area of multi-objective OCO by formulating the {\it min-max} static regret minimization problem, that requires an algorithm to 
closely track multiple loss function sequences simultaneously and at all times. Because of the lower bound on regret that scales linearly with the time horizon under an adversarial input,  we considered the i.i.d. input setting, for which we proposed a simple algorithm that used a combination of the well-known {\it Hedge} and OGD algorithm. Using an insightful regret decomposition, which made the analysis modular, we derived an $O(\sqrt{T\log (TK)})$ regret bound on the proposed algorithm. We also extended the results to the bandit information setting.   Certain questions that remain open are: i) how to handle the case when all loss functions are strongly convex, and ii) handling constraints like COCO.

We also generalized  the classical problem of {\it universal prediction of binary sequences} to the case when there are multiple sequences. 
Even for this binary case, we showed that the {\it min-max} static regret of any online algorithm scales linearly with the time horizon, exposing the 
inherent difficulty of the {\it min-max} setting. With stochastic input where $(b_t^1, b_t^2)$ are i.i.d. across time, we proposed an algorithm that chooses its action on the strength of statistical evidence and showed that it has $O(1)$ expected {\it min-max} static regret. 


\newpage
  \bibliographystyle{unsrt}
  \bibliography{OCO}

\begin{thebibliography}{53}
\providecommand{\natexlab}[1]{#1}
\providecommand{\url}[1]{\texttt{#1}}
\expandafter\ifx\csname urlstyle\endcsname\relax
  \providecommand{\doi}[1]{doi: #1}\else
  \providecommand{\doi}{doi: \begingroup \urlstyle{rm}\Url}\fi

\bibitem[Agarwal et~al.(2010)Agarwal, Hazan, Kale, and
  Schapire]{agarwal2010bandit}
Alekh Agarwal, Elad Hazan, Satyen Kale, and Robert~E. Schapire.
\newblock Optimal algorithms for online convex optimization with multi-point
  bandit feedback.
\newblock In \emph{Proceedings of the 23rd Annual Conference on Learning Theory
  (COLT)}, pages 28--40. Omnipress, 2010.

\bibitem[Azar et~al.(2014)Azar, Felge, Feldman, and
  Tennenholtz]{azar2014sequential}
Yossi Azar, Uriel Felge, Michal Feldman, and Moshe Tennenholtz.
\newblock Sequential decision making with vector outcomes.
\newblock In \emph{Proceedings of the 5th conference on Innovations in
  theoretical computer science}, pages 195--206, 2014.

\bibitem[Bertsimas et~al.(2011)Bertsimas, Brown, and
  Caramanis]{bertsimas2011theory}
Dimitris Bertsimas, David~B Brown, and Constantine Caramanis.
\newblock Theory and applications of robust optimization.
\newblock \emph{SIAM review}, 53\penalty0 (3):\penalty0 464--501, 2011.

\bibitem[Cao and Liu(2018)]{cao2018online}
Xuanyu Cao and KJ~Ray Liu.
\newblock Online convex optimization with time-varying constraints and bandit
  feedback.
\newblock \emph{IEEE Transactions on automatic control}, 64\penalty0
  (7):\penalty0 2665--2680, 2018.

\bibitem[Cardoso et~al.(2019)Cardoso, Abernethy, Wang, and
  Xu]{pmlr-v97-cardoso19a}
Adrian~Rivera Cardoso, Jacob Abernethy, He~Wang, and Huan Xu.
\newblock Competing against {N}ash equilibria in adversarially changing
  zero-sum games.
\newblock In Kamalika Chaudhuri and Ruslan Salakhutdinov, editors,
  \emph{Proceedings of the 36th International Conference on Machine Learning},
  volume~97 of \emph{Proceedings of Machine Learning Research}, pages 921--930.
  PMLR, 09--15 Jun 2019.
\newblock URL \url{https://proceedings.mlr.press/v97/cardoso19a.html}.

\bibitem[Cesa-Bianchi and Lugosi(2006)]{cesa2006prediction}
Nicolo Cesa-Bianchi and G{\'a}bor Lugosi.
\newblock \emph{Prediction, learning, and games}.
\newblock Cambridge university press, 2006.

\bibitem[Chen et~al.(2020)Chen, Zhang, and Weng]{chen2020practical}
Lijie Chen, Hongyang Zhang, and Tsui-Wei Weng.
\newblock Practical minimax fair classification.
\newblock In \emph{Advances in Neural Information Processing Systems
  (NeurIPS)}, 2020.

\bibitem[Chen and Giannakis(2018)]{chen2018bandit}
Tianyi Chen and Georgios~B Giannakis.
\newblock Bandit convex optimization for scalable and dynamic iot management.
\newblock \emph{IEEE Internet of Things Journal}, 6\penalty0 (1):\penalty0
  1276--1286, 2018.

\bibitem[Chouldechova and Roth(2020)]{chouldechova2020frontiers}
Alexandra Chouldechova and Aaron Roth.
\newblock The frontiers of fairness in machine learning.
\newblock \emph{Communications of the ACM}, 63\penalty0 (5):\penalty0 82--89,
  2020.

\bibitem[Chow et~al.(2018)Chow, Nachum, Duenez-Guzman, and
  Ghavamzadeh]{chow2018lyapunov}
Yinlam Chow, Ofir Nachum, Edgar Duenez-Guzman, and Mohammad Ghavamzadeh.
\newblock A lyapunov-based approach to safe reinforcement learning.
\newblock In \emph{Advances in Neural Information Processing Systems
  (NeurIPS)}, volume~31, 2018.

\bibitem[Cover(1966)]{cover1966behavior}
Thomas~M Cover.
\newblock \emph{Behavior of sequential predictors of binary sequences}.
\newblock Number 7002. Stanford University, Stanford Electronics Laboratories,
  Systems Theory~…, 1966.

\bibitem[Das and Banerjee(2011)]{das2011meta}
Sanmay Das and Arindam Banerjee.
\newblock Meta-optimization and multi-objective online portfolio selection.
\newblock In \emph{Proceedings of the Twenty-Fifth AAAI Conference on
  Artificial Intelligence (AAAI)}, pages 1100--1105, 2011.

\bibitem[Duchi et~al.(2015)Duchi, Jordan, Wainwright, and
  Wibisono]{duchi2015optimal}
John~C Duchi, Michael~I Jordan, Martin~J Wainwright, and Andre Wibisono.
\newblock Optimal rates for zero-order convex optimization: The power of two
  function evaluations.
\newblock \emph{IEEE Transactions on Information Theory}, 61\penalty0
  (5):\penalty0 2788--2806, 2015.

\bibitem[Even-Dar et~al.(2009)Even-Dar, Kleinberg, Mannor, and
  Mansour]{MannorGlobal}
Eyal Even-Dar, Robert Kleinberg, Shie Mannor, and Yishay Mansour.
\newblock Online learning with global cost functions.
\newblock In \emph{22nd Annual Conference on Learning Theory, COLT}, 2009.
\newblock URL \url{http://www.cs.mcgill.ca/~colt2009/papers/005.pdf#page=1}.

\bibitem[Feder et~al.(2002)Feder, Merhav, and Gutman]{feder2002universal}
Meir Feder, Neri Merhav, and Michael Gutman.
\newblock Universal prediction of individual sequences.
\newblock \emph{IEEE transactions on Information Theory}, 38\penalty0
  (4):\penalty0 1258--1270, 2002.

\bibitem[Flaxman et~al.(2005)Flaxman, Kalai, and McMahan]{flaxman2005bandit}
Abraham~D. Flaxman, Adam Kalai, and H.~Brendan McMahan.
\newblock Online convex optimization in the bandit setting.
\newblock In \emph{Proceedings of the 16th Annual Conference on Learning Theory
  (COLT)}, pages 206--211. Springer, 2005.

\bibitem[Freedman(1975)]{freedman1975tail}
David~A. Freedman.
\newblock On tail probabilities for martingales.
\newblock \emph{The Annals of Probability}, 3\penalty0 (1):\penalty0 100--118,
  1975.

\bibitem[Guo et~al.(2022)Guo, Liu, Wei, and Ying]{guo2022online}
Hengquan Guo, Xin Liu, Honghao Wei, and Lei Ying.
\newblock Online convex optimization with hard constraints: Towards the best of
  two worlds and beyond.
\newblock \emph{Advances in Neural Information Processing Systems},
  35:\penalty0 36426--36439, 2022.

\bibitem[Hao et~al.(2016)Hao, Kodialam, Lakshman, and Mukherjee]{hao2016online}
Fang Hao, Murali Kodialam, TV~Lakshman, and Sarit Mukherjee.
\newblock Online allocation of virtual machines in a distributed cloud.
\newblock \emph{IEEE/ACM Transactions on Networking}, 25\penalty0 (1):\penalty0
  238--249, 2016.

\bibitem[Hazan(2019)]{HazanBook}
Elad Hazan.
\newblock Introduction to online convex optimization.
\newblock \emph{CoRR}, abs/1909.05207, 2019.
\newblock URL \url{http://arxiv.org/abs/1909.05207}.

\bibitem[Jenatton et~al.(2016)Jenatton, Huang, and
  Archambeau]{jenatton2016adaptive}
Rodolphe Jenatton, Jim Huang, and C{\'e}dric Archambeau.
\newblock Adaptive algorithms for online convex optimization with long-term
  constraints.
\newblock In \emph{International Conference on Machine Learning}, pages
  402--411. PMLR, 2016.

\bibitem[Jiang et~al.(2023)Jiang, Zhang, Zhou, Gu, Zeng, and
  Zhu]{jiang2023multiobjective}
Jiyan Jiang, Wenpeng Zhang, Shiji Zhou, Lihong Gu, Xiaodong Zeng, and Wenwu
  Zhu.
\newblock Multi-objective online learning.
\newblock In \emph{The Eleventh International Conference on Learning
  Representations}, 2023.
\newblock URL \url{https://openreview.net/forum?id=dKkMnCWfVmm}.

\bibitem[Kesselheim and Singla(2020)]{kesselheim2020online}
Thomas Kesselheim and Sahil Singla.
\newblock Online learning with vector costs and bandits with knapsacks.
\newblock In \emph{Conference on Learning Theory}, pages 2286--2305. PMLR,
  2020.

\bibitem[Kouvelis and Yu(1997)]{kouvelis1997robust}
Panos Kouvelis and Gang Yu.
\newblock \emph{Robust Discrete Optimization and Its Applications}.
\newblock Springer, 1997.

\bibitem[Lee et~al.(2022)Lee, Noarov, Pai, and Roth]{lee2022online}
Daniel Lee, Georgy Noarov, Mallesh Pai, and Aaron Roth.
\newblock Online minimax multiobjective optimization: Multicalibeating and
  other applications.
\newblock \emph{Advances in Neural Information Processing Systems},
  35:\penalty0 29051--29063, 2022.

\bibitem[Liakopoulos et~al.(2019)Liakopoulos, Destounis, Paschos, Spyropoulos,
  and Mertikopoulos]{georgios-cautious}
Nikolaos Liakopoulos, Apostolos Destounis, Georgios Paschos, Thrasyvoulos
  Spyropoulos, and Panayotis Mertikopoulos.
\newblock Cautious regret minimization: Online optimization with long-term
  budget constraints.
\newblock In \emph{International Conference on Machine Learning}, pages
  3944--3952. PMLR, 2019.

\bibitem[Lin et~al.(2012)Lin, Wierman, Andrew, and Thereska]{lin2012dynamic}
Minghong Lin, Adam Wierman, Lachlan~LH Andrew, and Eno Thereska.
\newblock Dynamic right-sizing for power-proportional data centers.
\newblock \emph{IEEE/ACM Transactions on Networking}, 21\penalty0 (5):\penalty0
  1378--1391, 2012.

\bibitem[Liu and Hajiesmaili(2025)]{Hajiesmailifair}
Qingsong Liu and Mohammad Hajiesmaili.
\newblock Online fair allocation of reusable resources.
\newblock \emph{Proc. ACM Meas. Anal. Comput. Syst.}, 9\penalty0 (2), June
  2025.
\newblock \doi{10.1145/3727121}.
\newblock URL \url{https://doi.org/10.1145/3727121}.

\bibitem[Liu et~al.(2022)Liu, Wu, Huang, and Fang]{liu2022simultaneously}
Qingsong Liu, Wenfei Wu, Longbo Huang, and Zhixuan Fang.
\newblock Simultaneously achieving sublinear regret and constraint violations
  for online convex optimization with time-varying constraints.
\newblock \emph{ACM SIGMETRICS Performance Evaluation Review}, 49\penalty0
  (3):\penalty0 4--5, 2022.

\bibitem[Lyu et~al.(2019)Lyu, Cheung, Teo, and Wang]{lyu2019multi}
Guodong Lyu, Wang~Chi Cheung, Chung-Piaw Teo, and Hai Wang.
\newblock Multi-objective online ride-matching.
\newblock \emph{Available at SSRN}, 3356823:\penalty0 1--41, 2019.

\bibitem[Mahdavi et~al.(2012)Mahdavi, Jin, and Yang]{mahdavi2012trading}
Mehrdad Mahdavi, Rong Jin, and Tianbao Yang.
\newblock Trading regret for efficiency: online convex optimization with long
  term constraints.
\newblock \emph{The Journal of Machine Learning Research}, 13\penalty0
  (1):\penalty0 2503--2528, 2012.

\bibitem[Mahdavi et~al.(2013)Mahdavi, Yang, and Jin]{mahdavi2013stochastic}
Mehrdad Mahdavi, Tianbao Yang, and Rong Jin.
\newblock Stochastic convex optimization with multiple objectives.
\newblock \emph{Advances in neural information processing systems}, 26, 2013.

\bibitem[Milan et~al.(2017)Milan, Rezatofighi, Dick, Reid, and
  Schindler]{targettracking}
Anton Milan, S~Hamid Rezatofighi, Anthony Dick, Ian Reid, and Konrad Schindler.
\newblock Online multi-target tracking using recurrent neural networks.
\newblock In \emph{Proceedings of the AAAI conference on Artificial
  Intelligence}, volume~31, 2017.

\bibitem[Neely(2010)]{neely2010stochastic}
Michael~J Neely.
\newblock Stochastic network optimization with application to communication and
  queueing systems.
\newblock \emph{Synthesis Lectures on Communication Networks}, 3\penalty0
  (1):\penalty0 1--211, 2010.

\bibitem[Neely and Yu(2017)]{neely2017online}
Michael~J Neely and Hao Yu.
\newblock Online convex optimization with time-varying constraints.
\newblock \emph{arXiv preprint arXiv:1702.04783}, 2017.

\bibitem[Paulin(2015)]{paulin2015concentration}
Daniel Paulin.
\newblock Concentration inequalities for markov chains by marton couplings and
  spectral methods.
\newblock \emph{Electronic Journal of Probability}, 20:\penalty0 1--32, 2015.

\bibitem[Rakhlin et~al.(2011)Rakhlin, Sridharan, and Tewari]{rakhlinglobal}
Alexander Rakhlin, Karthik Sridharan, and Ambuj Tewari.
\newblock Online learning: Beyond regret.
\newblock In Sham~M. Kakade and Ulrike von Luxburg, editors, \emph{Proceedings
  of the 24th Annual Conference on Learning Theory}, volume~19 of
  \emph{Proceedings of Machine Learning Research}, pages 559--594, Budapest,
  Hungary, 09--11 Jun 2011. PMLR.
\newblock URL \url{https://proceedings.mlr.press/v19/rakhlin11a.html}.

\bibitem[Sener and Koltun(2018)]{multitask}
Ozan Sener and Vladlen Koltun.
\newblock Multi-task learning as multi-objective optimization.
\newblock In S.~Bengio, H.~Wallach, H.~Larochelle, K.~Grauman, N.~Cesa-Bianchi,
  and R.~Garnett, editors, \emph{Advances in Neural Information Processing
  Systems}, volume~31. Curran Associates, Inc., 2018.
\newblock URL
  \url{https://proceedings.neurips.cc/paper_files/paper/2018/file/432aca3a1e345e339f35a30c8f65edce-Paper.pdf}.

\bibitem[Shamir(2017)]{shamir2017optimal}
Ohad Shamir.
\newblock An optimal algorithm for bandit and zero-order convex optimization
  with two-point feedback.
\newblock \emph{Journal of Machine Learning Research}, 18\penalty0
  (52):\penalty0 1--11, 2017.

\bibitem[Si~Salem et~al.(2023)Si~Salem, Iosifidis, and Neglia]{si2023enabling}
Tarek Si~Salem, George Iosifidis, and Giovanni Neglia.
\newblock Enabling long-term fairness in dynamic resource allocation.
\newblock In \emph{Proceedings of the ACM SIGMETRICS International Conference
  on Measurement and Modeling of Computer Systems}, New York, NY, USA, 2023.
  ACM.

\bibitem[Sinha and Vaze(2024)]{Sinha2024}
Abhishek Sinha and Rahul Vaze.
\newblock Optimal algorithms for online convex optimization with adversarial
  constraints.
\newblock In \emph{The Thirty-eighth Annual Conference on Neural Information
  Processing Systems}, 2024.
\newblock URL \url{https://openreview.net/forum?id=TxffvJMnBy}.

\bibitem[Sinha and Vaze(2025)]{Vaze2025b}
Abhishek Sinha and Rahul Vaze.
\newblock Beyond \${\textbackslash}tilde\{O\}({\textbackslash}sqrt\{T\})\$
  constraint violation for online convex optimization with adversarial
  constraints.
\newblock In \emph{The Thirty-ninth Annual Conference on Neural Information
  Processing Systems, NeurIPS 2025}, 2025.
\newblock URL \url{https://openreview.net/forum?id=yK4Xu7DDd6}.

\bibitem[Sinha et~al.(2023)Sinha, Joshi, Bhattacharjee, Musco, and
  Hajiesmaili]{abhishekFair}
Abhishek Sinha, Ativ Joshi, Rajarshi Bhattacharjee, Cameron Musco, and Mohammad
  Hajiesmaili.
\newblock No-regret algorithms for fair resource allocation.
\newblock In A.~Oh, T.~Naumann, A.~Globerson, K.~Saenko, M.~Hardt, and
  S.~Levine, editors, \emph{Advances in Neural Information Processing Systems},
  volume~36, pages 48083--48109. Curran Associates, Inc., 2023.
\newblock URL
  \url{https://proceedings.neurips.cc/paper_files/paper/2023/file/96842011407c2691ab4eefff48fc864d-Paper-Conference.pdf}.

\bibitem[Sun et~al.(2017)Sun, Dey, and Kapoor]{pmlr-v70-sun17a}
Wen Sun, Debadeepta Dey, and Ashish Kapoor.
\newblock Safety-aware algorithms for adversarial contextual bandit.
\newblock In \emph{International Conference on Machine Learning}, pages
  3280--3288. PMLR, 2017.

\bibitem[Urgaonkar et~al.(2010)Urgaonkar, Kozat, Igarashi, and
  Neely]{NeelyFair}
Rahul Urgaonkar, Ulas~C. Kozat, Ken Igarashi, and Michael~J. Neely.
\newblock Dynamic resource allocation and power management in virtualized data
  centers.
\newblock In \emph{2010 IEEE Network Operations and Management Symposium - NOMS
  2010}, pages 479--486, 2010.
\newblock \doi{10.1109/NOMS.2010.5488484}.

\bibitem[Uziel and El-Yaniv(2017)]{uziel2017multi}
Guy Uziel and Ran El-Yaniv.
\newblock Multi-objective non-parametric sequential prediction.
\newblock \emph{Advances in Neural Information Processing Systems}, 30, 2017.

\bibitem[Vaze(2022)]{vazecocowiopt2022}
Rahul Vaze.
\newblock On dynamic regret and constraint violations in constrained online
  convex optimization.
\newblock In \emph{2022 20th International Symposium on Modeling and
  Optimization in Mobile, Ad hoc, and Wireless Networks (WiOpt)}, pages 9--16,
  2022.
\newblock \doi{10.23919/WiOpt56218.2022.9930613}.

\bibitem[Vaze and Sinha(2025)]{Vaze2025a}
Rahul Vaze and Abhishek Sinha.
\newblock \$o({\textbackslash}sqrt\{T\})\$ static regret and instance dependent
  constraint violation for constrained online convex optimization.
\newblock In \emph{The Thirty-ninth Annual Conference on Neural Information
  Processing Systems, NeurIPS 2025}, 2025.
\newblock URL \url{https://openreview.net/forum?id=YmbQ0qnQ76}.

\bibitem[Vershynin(2018)]{vershynin2018high}
Roman Vershynin.
\newblock \emph{High-Dimensional Probability: An Introduction with Applications
  in Data Science}.
\newblock Cambridge Series in Statistical and Probabilistic Mathematics.
  Cambridge University Press, 2018.
\newblock \doi{10.1017/9781108231596}.
\newblock Lemma 5.2: Discretization via $\varepsilon$-nets for Lipschitz
  functions.

\bibitem[Yi et~al.(2021)Yi, Li, Yang, Xie, Chai, and Johansson]{yi2021regret}
Xinlei Yi, Xiuxian Li, Tao Yang, Lihua Xie, Tianyou Chai, and Karl Johansson.
\newblock Regret and cumulative constraint violation analysis for online convex
  optimization with long term constraints.
\newblock In \emph{International Conference on Machine Learning}, pages
  11998--12008. PMLR, 2021.

\bibitem[Yi et~al.(2023)Yi, Li, Yang, Xie, Hong, Chai, and
  Johansson]{yi2023distributed}
Xinlei Yi, Xiuxian Li, Tao Yang, Lihua Xie, Yiguang Hong, Tianyou Chai, and
  Karl~H Johansson.
\newblock Distributed online convex optimization with adversarial constraints:
  Reduced cumulative constraint violation bounds under slater's condition.
\newblock \emph{arXiv preprint arXiv:2306.00149}, 2023.

\bibitem[Yu et~al.(2017)Yu, Neely, and Wei]{yu2017online}
Hao Yu, Michael Neely, and Xiaohan Wei.
\newblock Online convex optimization with stochastic constraints.
\newblock \emph{Advances in Neural Information Processing Systems}, 30, 2017.

\bibitem[Yuan and Lamperski(2018)]{yuan2018online}
Jianjun Yuan and Andrew Lamperski.
\newblock Online convex optimization for cumulative constraints.
\newblock \emph{Advances in Neural Information Processing Systems}, 31, 2018.

\end{thebibliography}
  \newpage
  \section{Experimental Results}\label{sec:sim}
In this section, we present numerical results on the true {\it min-max} static regret of Algorithm \ref{alg:main} under relevant settings, that are well motivated from both theoretical as well applications point of view. In all the figures, we plot the true {\it min-max} static regret as a function of time horizon $T$.

\subsection{Basic Results}
\begin{figure}[h!]
    \centering
    \begin{tikzpicture}
    \begin{axis}[
    width=.82\textwidth, 
        height=7cm, 
        title={True {\it min-max} static regret for random linear functions},
        xlabel={Time ($t$)},
        ylabel={Average Cumulative Regret},
        xmin=0, xmax=310, 
        ymin=0, ymax=75,  
        grid=major,
        minor tick num=1,
        major grid style={dotted, gray!50},
        legend pos=south east,
        legend style={
            fill=white, 
            opacity=0.9, 
            draw=gray, 
            font=\small
        },
        thick,
        cycle list name=color, 
    ]

    \addplot[mark=*, mark options={fill=white}, color=blue] 
        coordinates {
            (10, 10.0) (50, 20.0) (100, 35.0) (200, 45.0) (300, 53.4327)
        };
    \addlegendentry{$K=1$}

    \addplot[mark=square*, mark options={fill=white}, color=red!70!black] 
        coordinates {
            (10, 15.0) (50, 30.0) (100, 40.0) (200, 50.0) (300, 59.9533)
        };
    \addlegendentry{$K=10$}

    \addplot[mark=triangle*, mark options={fill=white}, color=green!60!black] 
        coordinates {
            (10, 20.0) (50, 40.0) (100, 50.0) (200, 60.0) (300, 66.2988)
        };
    \addlegendentry{$K=50$}
    
    \addplot[mark=o, color=orange] 
        coordinates {
            (10, 22.0) (50, 42.0) (100, 52.0) (200, 62.0) (300, 67.6223)
        };
    \addlegendentry{$K=100$}

    \end{axis}
    \end{tikzpicture}
    \caption{True {\it min-max} static regret of Algorithm \ref{alg:main} when $f_t^k$ are random linear functions for different 
    values of  $K$.}
    \label{fig:basic}
\end{figure}
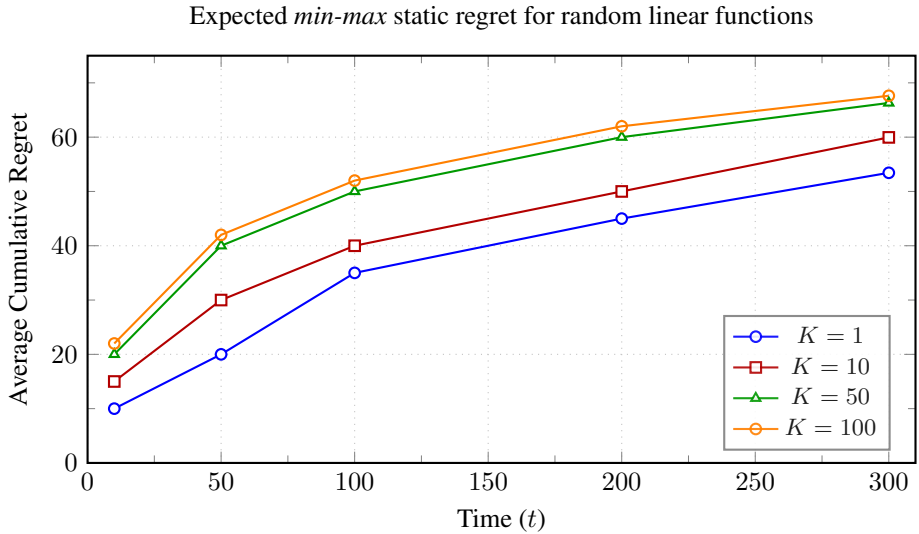

We begin our experimental results by considering the input model when all functions $f_t^k$ for $k=1,\dots, K, t=1, \dots, T$ are 
random linear functions in $d=10$ dimensions, where each coefficient is distributed uniformly between $[0,1]$.  Random linear functions are important  since they were sufficient to find lower bounds on the regret of any online algorithm for OCO \cite{HazanBook}.
In Fig. \ref{fig:basic}, we plot the true {\it min-max} static regret of Algorithm \ref{alg:main} as a function of $K$ and $T$ for random linear functions, where,  as expected, we observe that the regret scales as $O(\sqrt{T \log K})$. 

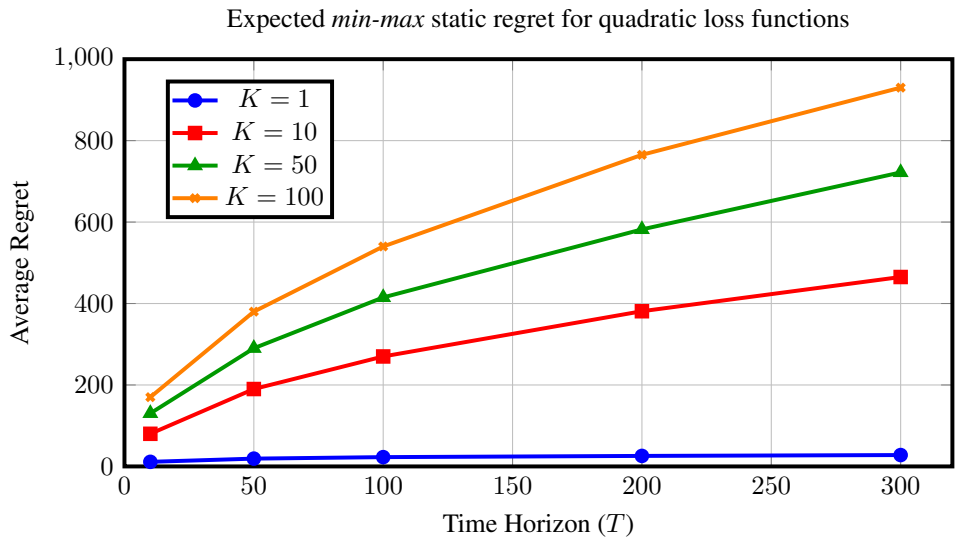
\begin{figure}
\begin{tikzpicture}
    \begin{axis}[
     width=.82\textwidth, 
        height=7cm,
        xlabel={Time Horizon ($T$)},
        ylabel={Average Regret},
        title={True {\it min-max} static regret for quadratic loss functions},
        xmin=0, xmax=320,
        ymin=0, ymax=1000, 
        grid=major,
        legend style={at={(0.05,0.95)}, anchor=north west, legend columns=1, inner sep=1pt, nodes={inner sep=2pt, minimum height=0pt}},
        line width=1.5pt,
    ]
    \addplot[blue, mark=*,] coordinates {
        (10, 11.27)
        (50, 19.16)
        (100, 22.84)
        (200, 25.81)
        (300, 27.93)
    };
    \addlegendentry{$K=1$}
    \addplot[red, mark=square*] coordinates {
        (10, 80.12)
        (50, 190.05)
        (100, 270.01)
        (200, 381.08)
        (300, 465.11)
    };
    \addlegendentry{$K=10$}
    \addplot[green!60!black, mark=triangle*] coordinates {
        (10, 130.34)
        (50, 290.18)
        (100, 415.02)
        (200, 582.01)
        (300, 721.90)
    };
    \addlegendentry{$K=50$}
    \addplot[orange, mark=x] coordinates {
        (10, 170.04)
        (50, 380.01)
        (100, 540.09)
        (200, 765.11)
        (300, 930.08)
    };
    \addlegendentry{$K=100$}
    \end{axis}
\end{tikzpicture}
\caption{True {\it min-max} static regret of Algorithm \ref{alg:main} when $f_t^k$ are random quadratic functions for different 
    values of  $K$.}\label{fig:quad}
\end{figure}

Next, we consider $f_t^k$'s as strongly convex functions, where each $f_t^k$ is independently and identically selected from a set of 
$\{(x-a)^2, a = [-9, -8 , \dots, 0,1, \dots, 9]\}$. Recall that for OCO with strongly convex functions, the step size $\eta$ needed for OGD to get a regret guarantee of $O(\log T)$ scales as $\eta=1/t$ and not $\eta=1/\sqrt{t}$   \cite{HazanBook}. Thus, for this simulation of Algorithm \ref{alg:main}\footnote{In contrast to $\eta_{x,t} = O(1/\sqrt{t})$ as needed to prove Theorem \ref{thm:main}.} we use $\eta_{\theta,t} = O(1/\sqrt{t})$ and $\eta_{x,t} = O(1/t)$. \footnote{We have checked that other combinations only increase the regret.}
As pointed out earlier, we are unable to improve the true {\it min-max} static regret guarantee for Algorithm \ref{alg:main} when all functions $f_t^k$'s are strongly convex. We see that in fact that is true for Algorithm \ref{alg:main} even in simulations as shown in Fig. \ref{fig:quad}, where for $K=1$, which is the same as OCO (with $\eta_{x,t} = O(1/t)$) the regret scales as $\log (T)$, while for larger values of $K$, the true {\it min-max} static regret scales as $O(\sqrt{T})$ when $\eta_{\theta,t} = O(1/\sqrt{t})$ and $\eta_{x,t} = O(1/t)$. Thus, if indeed true {\it min-max} static regret of $O(\log T)$ is achievable then Algorithm \ref{alg:main} needs to be conceptually modified.

\subsection{GlobalExpertsProblem}
 Recall that 
\cite{MannorGlobal} 
derived a {\it min-max} static regret \eqref{intro-moregret-def} guarantee of 
$O(\sqrt{T}\log K)$ with adversarial input for algorithm \textsf{MULTI}: 
For $K=2$, $$x_1(t+1) = x_1(t) + \frac{x_2(t) \ell_2(t) - x_1(t) \ell_1(t)}{\sqrt{T}},$$ and $x_2(t) = 1-x_1(t)$. 
For general $K$, the $K=2$ algorithm is used recursively.
Compared to our Algorithm \ref{alg:main}, the \textsf{MULTI} algorithm \cite{MannorGlobal} is  very compute intensive  since it recursively uses an algorithm for $K=2$ and requires the knowledge of time horizon $T$.

In this section, we compare the performance of Algorithm \ref{alg:main} and  \textsf{MULTI} \cite{MannorGlobal} for the 
GlobalExpertsProblem with the i.i.d. input. For that purpose, we consider that the loss of each expert at time $t$ is drawn independently from a uniform distribution:
\[
\ell_{t,k} \stackrel{\text{i.i.d.}}{\sim} \mathsf{Uniform}(a,b), \quad
t = 1,2,\dots,T,\; k = 1,2,\dots,K,
\]
where $a,b \in [0,1]$ are fixed parameters defining the range of losses.  
In our experiments, we set $a=0.2$ and $b=0.8$ to ensure non-trivial loss variation while avoiding extreme loss values.

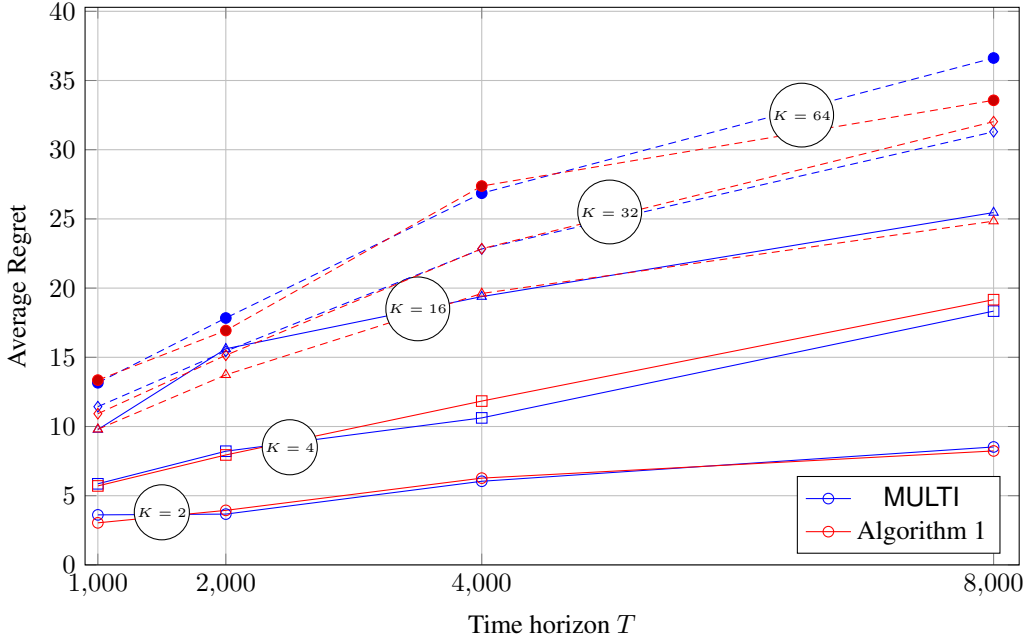
\begin{figure}
\centering
\begin{tikzpicture}
    \begin{axis}[
        width=14cm, height=9cm,
        xlabel={Time horizon $T$},
        ylabel={Average Regret},
        xmin=900, xmax=8200,
        ymin=0,
        xtick={1000,2000,4000,8000},
        yticklabel style={/pgf/number format/fixed},
        grid=major,
        legend style={at={(0.98,0.02)},anchor=south east},
        ymajorgrids=true,
        xmajorgrids=true,
        scaled y ticks=false
    ]

    \def\colorMULTI{blue}
    \def\colorA{red}

    \addplot+[color=\colorMULTI,mark=o] coordinates {
        (1000,3.6151) (2000,3.6701) (4000,6.0430) (8000,8.5220)
    };
    \addplot+[color=\colorA,mark=o] coordinates {
        (1000,3.0395) (2000,3.9405) (4000,6.2695) (8000,8.2332)
    };
    \node at (axis cs:1500,3.8) [circle,draw=black,fill=white,inner sep=0.8pt] {\tiny $K=2$};

    \addplot+[color=\colorMULTI,mark=square] coordinates {
        (1000,5.8597) (2000,8.2149) (4000,10.6216) (8000,18.3370)
    };
    \addplot+[color=\colorA,mark=square] coordinates {
        (1000,5.7061) (2000,7.9442) (4000,11.8377) (8000,19.1628)
    };
    \node at (axis cs:2500,8.5) [circle,draw=black,fill=white,inner sep=0.8pt] {\tiny $K=4$};

    \addplot+[color=\colorMULTI,mark=triangle] coordinates {
        (1000,9.7794) (2000,15.6170) (4000,19.3847) (8000,25.4587)
    };
    \addplot+[color=\colorA,mark=triangle] coordinates {
        (1000,9.7973) (2000,13.7320) (4000,19.6132) (8000,24.8382)
    };
    \node at (axis cs:3500,18.5) [circle,draw=black,fill=white,inner sep=0.8pt] {\tiny $K=16$}; 

    \addplot+[color=\colorMULTI,mark=diamond] coordinates {
        (1000,11.4337) (2000,15.4393) (4000,22.8256) (8000,31.2938)
    };
    \addplot+[color=\colorA,mark=diamond] coordinates {
        (1000,10.9234) (2000,15.1460) (4000,22.8466) (8000,32.0281)
    };
    \node at (axis cs:5000,25.5) [circle,draw=black,fill=white,inner sep=0.8pt] {\tiny $K=32$}; 

    \addplot+[color=\colorMULTI,mark=*] coordinates {
        (1000,13.1593) (2000,17.8398) (4000,26.8563) (8000,36.6256)
    };
    \addplot+[color=\colorA,mark=*] coordinates {
        (1000,13.3527) (2000,16.9349) (4000,27.3750) (8000,33.5698)
    };
    \node at (axis cs:6500,32.5) [circle,draw=black,fill=white,inner sep=0.8pt] {\tiny $K=64$}; 

    \addlegendimage{color=\colorMULTI,mark=*,only marks}
    \addlegendentry{\textsf{MULTI}}
    \addlegendimage{color=\colorA,mark=*,only marks}
    \addlegendentry{Algorithm \ref{alg:main}}
    \end{axis}
\end{tikzpicture}
\caption{ True {\it min-max} static regret of \textsf{MULTI} \cite{MannorGlobal} and Algorithm \ref{alg:main} for different numbers of experts $K$ and time horizons $T$ for the fair classification example.}
\label{fig:GEP}
\end{figure}

For this model, we plot the true {\it min-max} static regret performance of Algorithm \ref{alg:main} and  \textsf{MULTI} in Fig. \ref{fig:GEP}, and see that both perform similarly. It appears that \textsf{MULTI} has better performance than its guarantee \cite{MannorGlobal}.
However, 
\textsf{MULTI} is computationally far more intensive than Algorithm \ref{alg:main}.

\subsection{Application of {\it min-max} formulation to  Fair Classification Setting}\label{sec:simfairML}
Finally, in this subsection, we simulate the i.i.d.  {\it min-max} OCO problem for a fair classification setting.

\paragraph{Data generation.}
We consider $K$ groups, each corresponding to a convex loss function sequence.  
For each group $k \in [K]$, we sample a hidden parameter $w_k^\star \sim \mathcal{N}(0, I_d)$.  
At each round $t=1,\dots,T$, for each group $k$ we draw $m$ feature vectors
\[
z_{t,k,i} \sim \mathcal{N}(\mu_k, \sigma^2 I_d), \quad i=1,\dots,m,
\]
with $\mu_k$ drawn once per group. Labels are generated as
\[
y_{t,k,i} \sim \mathrm{Bernoulli}\!\left(\sigma(\langle w_k^\star, z_{t,k,i}\rangle)\right),
\]
where $\sigma(u) = 1/(1+e^{-u})$ is the sigmoid function.

\paragraph{Loss functions.}
The per-group convex loss at round $t$ is the averaged logistic loss with $L_2$ regularization:
\[
f_t^k(x) \;=\; \frac{1}{m}\sum_{i=1}^m \log\!\Big(1+\exp(-y_{t,k,i}\,\langle x,z_{t,k,i}\rangle)\Big) \;+\; \kappa \|x\|_2^2 .
\]

For this application, we compare the performance of following algorithms.
\begin{itemize}
    \item \textbf{Algorithm \ref{alg:main} {\it Hedge}+OGD})
        \item \textbf{OGD on average loss}: projected OGD on $\tfrac{1}{K}\sum_{k=1}^K f_t^k(x)$.
    \item \textbf{Follow-the-Regularized-Leader (FTRL)}: $x_t = \arg\min_{x \in \mathcal{X}} \left\{\max_k \sum_{s=1}^{t-1} f_s^k(x) + \cR(x)\right\}$.
\end{itemize}

\paragraph{Parameters.}
We use $d=20$, $K=10$, mini-batch size $m=50$, diameter $D=5$, and regularization $\kappa=10^{-3}$.  
Step sizes are set as $\eta_{x,t} = D/(G\sqrt{t})$ for OGD (with $G$ an upper bound on gradient norms) and 
$\eta_{\theta,t} = \sqrt{\ln K}/(B\sqrt{t})$ for {\it Hedge}, with $B$ an upper bound on per-round losses.

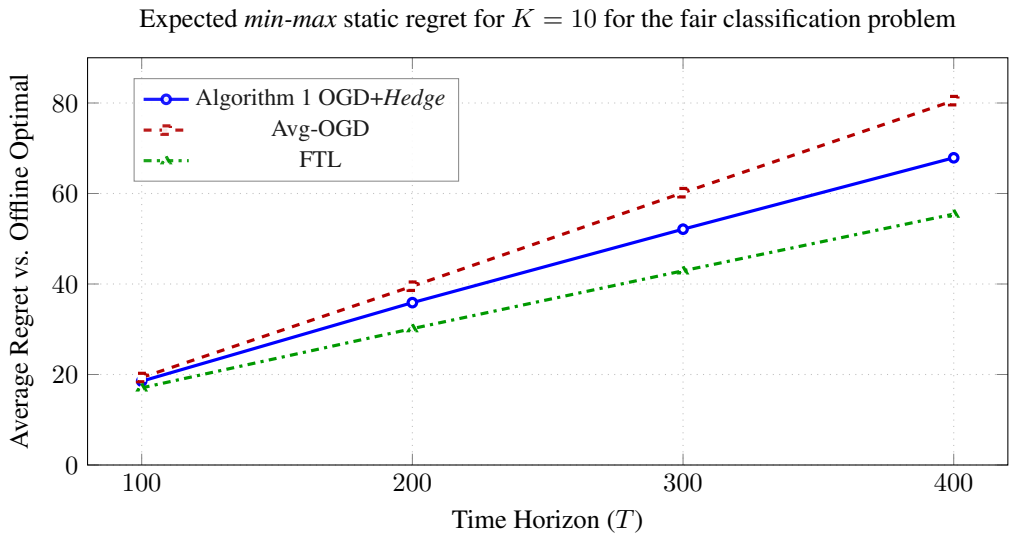
\begin{figure}[h!]
    \centering
    \begin{tikzpicture}
    \begin{axis}[
        width=0.9\textwidth, 
        height=7cm,        
        title={Expected {\it min-max} static regret for $K=10$ for the fair classification problem},
        xlabel={Time Horizon ($T$)},
        ylabel={Average Regret vs. Offline Optimal},
        xmin=80, xmax=420,  
        ymin=0, ymax=90,    
        grid=major,
        major grid style={dotted, gray!50},
        xtick={100, 200, 300, 400},       
        legend pos=north west,
        legend style={
            fill=white, 
            opacity=0.9, 
            draw=gray, 
            font=\small,
            at={(0.05, 0.95)} 
        },        
        every axis plot/.append style={very thick},
        mark options={scale=0.8, fill=white},
    ]
    \addplot[mark=*, color=blue] 
        coordinates {
            (100, 18.52) 
            (200, 35.88) 
            (300, 52.11) 
            (400, 67.89)
        };
    \addlegendentry{Algorithm \ref{alg:main} {\it Hedge}+OGD}

    \addplot[mark=square*, color=red!70!black, dashed] 
        coordinates {
            (100, 19.33) 
            (200, 39.51) 
            (300, 60.18) 
            (400, 80.52)
        };
    \addlegendentry{Avg-OGD}

    \addplot[mark=triangle*, color=green!60!black, dash dot] 
        coordinates {
            (100, 17.05) 
            (200, 30.14) 
            (300, 42.92) 
            (400, 55.45)
        };
    \addlegendentry{FTL}

    \end{axis}
    \end{tikzpicture}
    \caption{True {\it min-max} static regret comparison for $K=10$.}
    \label{fig:fairML}
\end{figure}

In Fig. \ref{fig:fairML}, we plot the expected {\it min-max} static regret ${\bar R}_T$ for all the three algorithms.
We see that all algorithms have sub-linear regret, and Algorithm \ref{alg:main} has larger regret than FTRL but better than average OGD.
 However, recall that neither FTRL and average OGD have any theoretical guarantees on their expected {\it min-max} static regret. Next, in Fig. \ref{fig:regret_switching}, we will consider an input where the index of the most challenging sequence changes over time and show that Algorithm \ref{alg:main} outperforms FTRL in that case.

\subsubsection*{Non-stationary input}
Next, we change the input described in Section \ref{sec:simfairML} to model  a \textbf{highly non‑stationary environment} specifically designed to challenge static optimization strategies such as FTRL and highlight the adaptability of Algorithm \ref{alg:main} that uses OGD+{\it Hedge}. The detailed changes with respect to Section \ref{sec:simfairML} are as follows:

\begin{itemize}
    \item We consider $K=3$ separate objectives, each with its own logistic loss function defined by fixed but distinct parameter vectors $w_\text{star}[k]$ and mean feature shifts $\mu[k]$.
    \item Every \texttt{switch\_interval} steps (here set to 100), one objective becomes significantly harder by adding a \textbf{large mean shift} ($\texttt{shift\_magnitude}=5.0$) to its feature distribution. This creates abrupt changes in which objective dominates the cumulative loss objective function.
    \item \textbf{Data generation:} For each round $t$, and for each objective $k$, a feature matrix $Z$ of size $m \times d$ is generated from a Gaussian distribution centered at the shifted mean. Labels $y$ are generated from a logistic model using $w_\text{star}[k]$ (is a hidden “true” parameter vector for objective $k$. It defines how that objective behaves.), with additional noise introduced via random sampling.
  \end{itemize}

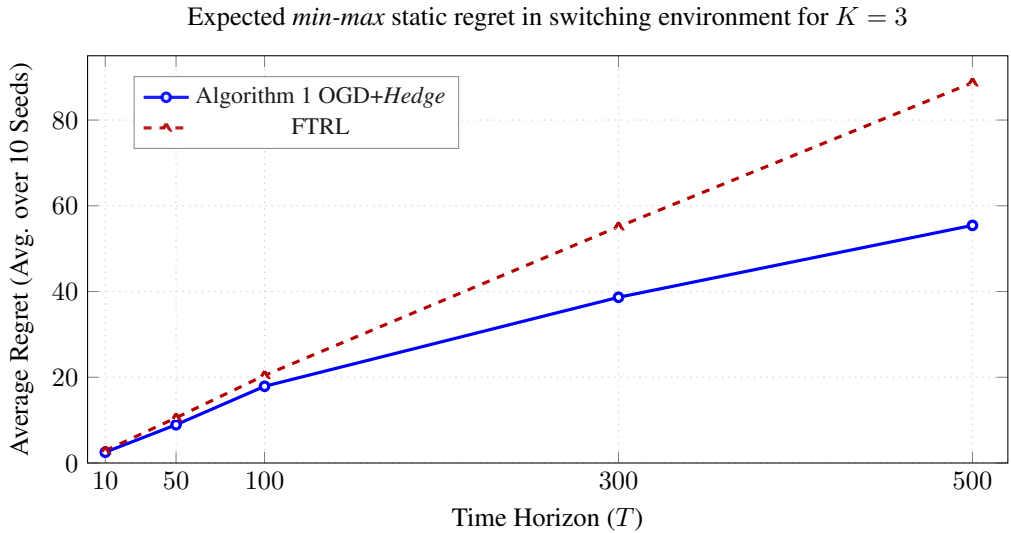
\begin{figure}[h!]
    \centering
    \begin{tikzpicture}
    \begin{axis}[
        width=0.9\textwidth, 
        height=7cm,        
        title={Expected {\it min-max} static regret in switching environment for $K=3$},
        xlabel={Time Horizon ($T$)},
        ylabel={Average Regret (Avg. over 10 Seeds)},
        xmin=0, xmax=520,  
        ymin=0, ymax=95,        
        grid=major,
        major grid style={dotted, gray!50},
        xtick={10, 50, 100, 300, 500},        
        legend pos=north west,
        legend style={
            fill=white, 
            opacity=0.9, 
            draw=gray, 
            font=\small,
            at={(0.05, 0.95)} 
        },        
        every axis plot/.append style={very thick},
        mark options={scale=0.8, fill=white},
    ]
    \addplot[mark=*, color=blue] 
        coordinates {
            (10, 2.51) 
            (50, 8.92) 
            (100, 17.89) 
            (300, 38.65) 
            (500, 55.43)
        };
    \addlegendentry{Algorithm \ref{alg:main} {\it Hedge}+OGD}
    \addplot[mark=triangle*, color=red!70!black, dashed] 
        coordinates {
            (10, 2.87) 
            (50, 10.55) 
            (100, 20.44) 
            (300, 55.12) 
            (500, 88.71)
        };
    \addlegendentry{FTRL}

    \end{axis}
    \end{tikzpicture}
    \caption{Expected {\it min-max} static regret for Algorithm \ref{alg:main}  vs. FTRL in a piecewise-stationary environment.}
    \label{fig:regret_switching}
\end{figure}

We clearly see in Fig. \ref{fig:regret_switching} that Algorithm \ref{alg:main} ({\it Hedge}+OGD) adapts better to changing conditions than FTRL, since worst‑case objective changes over time, making past loss information misleading for FTRL.

\appendix
\section{Applications of Multi-objective OCO and Open Questions}\label{app:openQ}
\subsection{Applications}
The {\it min-max} approach is also well suited for modern networked systems—such as data centers, wireless edge networks, and shared cloud infrastructures—where online decisions must balance performance across multiple service classes, users, or flows whose objectives (throughput, latency, energy) often conflict. The min-max regret criterion provides a principled way to ensure worst-case fairness by minimizing the largest cumulative loss among competing entities, which is crucial in fair resource allocation \cite{abhishekFair}, fair allocation in clouds \cite{hao2016online}, fair assignment of reusable resources \cite{Hajiesmailifair} congestion-controlled transport \cite{kouvelis1997robust, bertsimas2011theory}, and dynamic resource provisioning in data centers \cite{NeelyFair}. Unlike average-regret formulations that can conceal systematic bias, minimizing {\it min-max} regret explicitly guards against persistent under-allocation to any group or flow—an increasingly important property in multi-tenant and QoS-sensitive networks.

Multi-objective OCO has also been studied from an applications perspective, e.g. \cite{lyu2019multi} studied the problem of multi-period multi-objective online ride-matching problem, where the multiple objectives included platform revenue, assigning drivers with higher service quality, minimizing the distance to the customer etc. \cite{lyu2019multi} also used the 
COCO-type reformulation of multi-objective OCO for deriving theoretical guarantees. Other applications of multi-objective OCO include  multi-task learning \cite{multitask}, multiple target tracking \cite{targettracking}, portfolio selection \cite{das2011meta} and safe reinforcement learning \cite{chow2018lyapunov},  where decision-makers must make sequential decisions under uncertainty, balancing objectives such as cost, delivery time, and reliability without full knowledge of future demand or disruptions.

\subsection{Open Questions}
\begin{itemize}
\item Strongly Convex Functions:  For OCO, the best-known static regret reduces to $O(\log{T})$ when functions $f_t$'s are strongly convex.
Because of the use of  {\it Hedge} in our proposed algorithm to update weights $\theta_t$ , its expected {\it min-max} static regret \eqref{defn:expregretintro} remains $O(\sqrt{T})$ even if $f_t, g_t$ are strongly convex. 
It is not clear whether this is a limitation of the proposed algorithm or the expected {\it min-max} static regret is $\Omega(\sqrt{T})$ even when $f_t, g_t$ are strongly convex. Simulations (see Fig. \ref{fig:quad}) suggest that in fact expected {\it min-max} static regret is $\Omega(\sqrt{T})$ for the proposed algorithm.
\item Constrained Multi-Objective OCO: An immediate open question is how to solve the constrained version of the multi-objective OCO. For example, let $f_t, g_t$ be the loss functions under the ${\it min-max}$ criteria while $h_t$ be the constraint function. Similar to COCO, let $\mathcal{X}^\star$ be the feasible set consisting of all admissible actions that satisfy all constraints $h_{t}(x) \leq 0, t\in [T]$.
Then, the optimal expected static benchmark cost is 
\begin{equation}\label{intro:eq:optminmaxcoco}
{\bar C}_\opt = T\cdot \min_{x\in\cX^\star}\max_{\theta\in\Delta_2}\theta^\top\mu(x) \quad \text{with} \quad (x^\star, \theta^\star) = \arg \min_{x\in\cX^\star}\max_{\theta\in\Delta_2}\theta^\top\mu(x),
\end{equation}
and the corresponding regret and CCV are 
\begin{equation}\label{defn:expregretintrococo}
{\bar R}_{\cA}(T) = \sup_{\cD} \{{\bar C}_{\cA} -{\bar C}_\opt\} \quad \bar{\textrm{CCV}}_{[1:T]}  \equiv \sum_{t=1}^T \bbE\{ \max(h_{t}(x_t),0) \},
\end{equation} 
respectively. 
It turns out that our  proposed algorithm cannot be married readily with the prior best known approaches for COCO \cite{Sinha2024, Vaze2025a, Vaze2025b} for deriving simultaneous 
bounds on regret and CCV. Thus, how to solve the {\it min-max} OCO with constraints remains an interesting open problem.

\end{itemize}
\section{Related Work on COCO}\label{app:COCO}
Because of its widespread applications, multi-objective OCO has been an object of immense interest, however, with limited success.
For example, \cite{mahdavi2013stochastic} wanted to study the multi-objective OCO and stated the difficulty as ``{\it A fundamental
difference between single- and multi-objective optimization is that for the latter it is not
obvious how to evaluate the optimization quality}". Further, it said ``{\it Since it is impossible to simultaneously minimize
multiple loss functions and in order to avoid complications caused by handling more than
one objective, we choose one function as the objective and try to bound other objectives by appropriate
thresholds}. 
This work gave rise to the now popular problem of {\bf constrained online convex optimization} (COCO), where the problem is transformed into a constrained optimization problem with one objective and one or more constraint functions, that is described next. 

\subsubsection{COCO}
In COCO, on every round $t,$ an online algorithm first chooses an admissible action $x_t \in \mathcal{X} \subset \bbR^d,$ 
 and then the adversary chooses a convex loss/cost function $f_t: \mathcal{X} \to \mathbb{R}$ and a constraint function of the form $g_{t}(x) \leq 0,$ where $g_{t}: \mathcal{X} \to \mathbb{R}$ is a convex function. 
 Let $\mathcal{X}^\star$ be the feasible set consisting of all admissible actions that satisfy all constraints $g_{t}(x) \leq 0, t\in [T]$. A  standard assumption is made that $\mathcal{X}^\star$ is not empty (called the {\it feasibility assumption}).
 Since $g_{t}$'s are revealed after the action $x_t$ is chosen, an online algorithm $\cA$ need not take feasible actions on each round, and in addition to the static regret defined as 

 \begin{eqnarray} \label{intro-regret-def}
	\textrm{R}^{\text{COCO}}_\cA(T) \equiv \sup_{\{f_t\}_{t=1}^T} \left\{\sum_{t=1}^T f_t(x_t) -  \inf_{x \in \mathcal{X}^\star}\sum_{t=1}^T f_t(x^\star)\right\},
\end{eqnarray}
an additional metric of interest  is the total cumulative constraint violation (CCV)  defined as 
$$\textrm{CCV}_\cA(T)  \equiv \sum_{t=1}^T \max(g_{t}(x_t),0). $$
The goal with COCO is to design an online algorithm to simultaneously achieve a small regret \eqref{intro-regret-def} with respect to any admissible benchmark $x^\star \in \mathcal{X}^\star$ and a small CCV. 

\subsubsection{Prior Work on COCO}
 {\bf (A) i.i.d. loss and constraints:} COCO with independent and identically distributed (i.i.d.) loss functions $f_t$'s and constraint functions $g_t$'s was studied in \cite{mahdavi2013stochastic} while its extension to stationary and ergodic processes can be found in \cite{uziel2017multi}, where both derived  an algorithm with regret $O(\sqrt{T})$ and CCV $O(T^{3/4})$.
 
 {\bf  (B) Time-invariant constraints:} COCO with time-invariant constraints, \emph{i.e.,} $g_{t} = g, \forall \ t$ \citep{yuan2018online, jenatton2016adaptive, mahdavi2012trading, yi2021regret} has been considered extensively, where functions $g$ are assumed to be known to the algorithm \emph{a priori}.  The best known results in this case give static regret of $O(T^{\max\{c,1-c\}})$ and CCV of $O(T^{(1-c)/2})$ for any $c\in(0,1)$. The algorithm is allowed to take actions that are infeasible at any time to avoid the costly projection step of the vanilla projected OGD algorithm and the main objective was to design an \emph{efficient} algorithm  with a small regret and CCV while avoiding  the explicit projection step. 

{\bf (C) Time-varying constraints:} The more difficult question is solving COCO problem when the constraint functions, \emph{i.e.}, $g_{t}$'s, change arbitrarily with time $t$.  
One popular algorithm for solving COCO considered a Lagrangian function optimization that is updated using the primal and dual variables \citep{yu2017online, pmlr-v70-sun17a, yi2023distributed}. Alternatively, \citep{neely2017online} and \citep{georgios-cautious} used the drift-plus-penalty (DPP) framework  \citep{neely2010stochastic} to solve the COCO, but which needed additional assumption, e.g. the Slater's condition in \citep{neely2017online} and with weaker form of the feasibility assumption in \citep{neely2017online}. 

 \citep{guo2022online} obtained bounds similar to \citep{neely2017online} but without assuming Slater's condition. Very recently, the state of the art guarantees on simultaneous bounds on regret $O (\sqrt{T})$ and CCV $O (\sqrt{T}\log T)$ for COCO were derived in \citep{Sinha2024} with a very simple algorithm that combines the loss function at time $t$ and the CCV accrued till time $t$ in a single loss function, and then executes the online gradient descent (OGD) algorithm on the single loss function with an adaptive step-size. Most recently \citep{Vaze2025b} showed lower CCV is possible with increased regret compared to \citep{Sinha2024} while \citep{Vaze2025a} obtained instance dependent CCV bounds which can be $O(1)$ for specific cases while achieving regret of $O(\sqrt{T})$, respectively.
The COCO problem has been considered in the {\it dynamic} setting as well  \citep{chen2018bandit, cao2018online, vazecocowiopt2022, liu2022simultaneously} where the benchmark $x^\star$ in \eqref{intro-regret-def} is replaced by $x_t^\star$ ($x_t^\star = \arg \min_x f_t(x)$) that is also allowed to change its actions over time. 

\begin{remark}Even though the {\it min-max} OCO and the COCO problem appear related, but unfortunately results for one do not give results for the other in either direction.
\end{remark}
\section{Proof of Theorem \ref{thm:lb}}\label{app:lb}

\begin{proof}
Let the decision set be $\mathcal X=[-1,1]$. The adversary constructs two inputs $\sigma_1$ and $\sigma_2$ with common prefix until 
$t=T/2$ as follows.
For both inputs $\sigma_1$ and $\sigma_2$, for rounds $t=1,\dots,T/2$,
\[
f_t(x)=-x,
\qquad
g_t(x)=x.
\]

Let $x_t$ be the actions of an online algorithm $\cA$, and let
\[
{\bar x} := \sum_{t=1}^{T/2} x_t
\]
denote the cumulative action of $\cA$ during the prefix.
After observing ${\bar x}$, the adversary completes the input as follows:
\begin{itemize}
\item If ${\bar x}<0$, the adversary selects input $\sigma_1$.
\item If ${\bar x}\ge 0$, the adversary selects input $\sigma_2$, where the respective suffixes for $t=T/2+1,\dots, T$ are defined as follows.
\end{itemize}

\paragraph{Suffix of $\sigma_1$.}
For $t=T/2+1,\dots,T$,
\[
f_t(x)=1,
\qquad
g_t(x)=0.
\]

\paragraph{Suffix of $\sigma_2$.}
For $t=T/2+1,\dots,T$,
\[
f_t(x)=0,
\qquad
g_t(x)=1.
\]

For $\sigma_1$, the cumulative losses for a fixed $x\in[-1,1]$ are
\[
\sum_{t=1}^T f_t(x) = -\frac{T}{2}x + \frac{T}{2},
\qquad
\sum_{t=1}^T g_t(x) = \frac{T}{2}x.
\]
The two quantities are equal at $x=\tfrac12$, yielding
\[
C_\opt(\sigma_1)=\frac{T}{4}.
\]

Similarly, for $\sigma_2$,
\[
\sum_{t=1}^T f_t(x) = -\frac{T}{2}x,
\qquad
\sum_{t=1}^T g_t(x) = \frac{T}{2}x + \frac{T}{2},
\]
which are equal at $x=-\tfrac12$, giving
\[
C_\opt(\sigma_2)=\frac{T}{4}.
\]

Thus, in both cases,
\[
C_\opt(\sigma_i)=\frac{T}{4}, \qquad i=1,2.
\]

We now lower bound the cost incurred by $\mathcal A$.

\paragraph{Case 1: ${\bar x}<0$ (input $\sigma_1$).}
\[
\sum_{t=1}^T f_t(x_t)
=
-\sum_{t=1}^{T/2} x_t + \frac{T}{2}
=
\frac{T}{2}-{\bar x},
\]
\[
\sum_{t=1}^T g_t(x_t)
=
\sum_{t=1}^{T/2} x_t
=
{\bar x}.
\]
Hence,
\[
C_{\mathcal A}(\sigma_1)
=
\max\Bigl\{
\frac{T}{2}-{\bar x},\;
{\bar x}
\Bigr\} \ge \frac{T}{2},
\]
since ${\bar x}<0$.

\paragraph{Case 2: ${\bar x}\ge 0$ (input $\sigma_2$).}
\[
\sum_{t=1}^T f_t(x_t)
=
-\sum_{t=1}^{T/2} x_t
=
-{\bar x},
\]
\[
\sum_{t=1}^T g_t(x_t)
=
\sum_{t=1}^{T/2} x_t + \frac{T}{2}.
\]
Thus,
\[
C_{\mathcal A}(\sigma_2)
=
\max\Bigl\{
-{\bar x},\;
\frac{T}{2}+{\bar x}
\Bigr\}
\;\ge\;
\frac{T}{2},
\]
since ${\bar x}\ge 0$.

Thus, 

\[
\mathrm{Regret}_\cA(T)
=
C_{\mathcal A}(\sigma_i)-C_\opt(\sigma_i)
\ge
\frac{T}{4}
\]
for at least one $i=1,2$.
\end{proof}

\section{Connecting the two benchmarks $\bbE\{C_\opt\}$ and ${\bar C}_\opt$ \eqref{intro:eq:optminmax} \label{app:regretconnection}
}\label{app:benchmarkconnection}
Without loss of generality, let $\chi \subseteq [0,1]^d$. 
We will invoke Assumptions \ref{cvx} and \ref{bddness} throughout without repeatedly mentioning it.
Thus, for each $t=1,\dots,T$ and $x\in[0,1]^d$,  we have 
\[
\lambda_t(x)=(f_t(x),g_t(x))\in[0,B]^2,
\]
where $\{\lambda_t(\cdot)\}_{t=1}^T$ are i.i.d.\ across $t$.  
Since $f_t(x),g_t(x)$ is $G$-Lipschitz, each coordinate of $\lambda_t$ is $G$--Lipschitz in $x$.
Let the empirical  mean be
\[
\hat\mu_T(x)=\frac{1}{T}\sum_{t=1}^T \lambda_t(x),\] while as before
\[
\mu(x)=\mathbb{E}\{\lambda_t(x)\}.
\]
Let $\Phi
=\min_x\max_{\theta\in\Delta_2}\theta^\top \mu(x).$
Then from \eqref{defn:optcostintro} and \eqref{intro:eq:optminmax} \[
C_\opt
=\min_{x}\max\Big\{\sum_t f_t(x),\,\sum_t g_t(x)\Big\},
\qquad
{\bar C}_\opt
=T \Phi.
\]

Let $\phi(v):=\min_x\max_{\theta\in\Delta_2}\theta^\top v(x)
=\min_x\max\{v_1(x),v_2(x)\}$ for $v:[0,1]^d\to\mathbb{R}^2$.
\begin{lemma}[1--Lipschitz property of $\phi$]\label{lem:lipschitz}
For any $v,w:[0,1]^d\to\mathbb{R}^2$,
\[
\big|\phi(v)-\phi(w)\big|
\;\le\;
\sup_{x\in[0,1]^d}\ \|v(x)-w(x)\|_\infty.
\]
\end{lemma}

\begin{proof}
For each $x\in[0,1]^d$ define
\[
h_v(x)\;:=\;\max_{\theta\in\Delta_2}\theta^\top v(x)
\;=\;\|v(x)\|_\infty.
\]
Then $\phi(v)=\min_x h_v(x)$.

First, note the elementary inequality valid for any $a,b\in\mathbb{R}^2$:
\[
\max_{k} a_k - \max_{k} b_k
\;\le\;
\max_{k} (a_k-b_k)
\;=\;
\|a-b\|_\infty.
\tag{$\ast$}
\]
Using ($\ast$) pointwise with $a=v(x)$ and $b=w(x)$ gives
\[
h_v(x)-h_w(x)
\;=\;
\sup_{k} v_k(x) - \sup_{k} w_k(x)
\;\le\;
\sup_{k}\big(v_k(x)-w_k(x)\big)
\;=\;
\|v(x)-w(x)\|_\infty.
\]
Thus, we get \[
\phi(v)-\phi(w)
\;=\;
\inf_x h_v(x) - \inf_x h_w(x)
\;\le\;
\sup_{x}\big(h_v(x)-h_w(x)\big)
\;\le\;
\sup_{x}\|v(x)-w(x)\|_\infty.
\]
Swapping the roles of $v$ and $w$ gives the reverse inequality
$\phi(w)-\phi(v)\le \sup_x\|v(x)-w(x)\|_\infty$.
Combining the two displays proves
\[
\big|\phi(v)-\phi(w)\big|
\;\le\;
\sup_{x\in[0,1]^d}\ \|v(x)-w(x)\|_\infty.
\]
\end{proof}

Hence, from Lemma \ref{lem:lipschitz}
\begin{equation}
\Big|\frac{C_\opt}{T}-\Phi\Big|
=|\phi(\hat\mu_T)-\phi(\mu)|
\le
\sup_{x\in[0,1]^d}\|\hat\mu_T(x)-\mu(x)\|_\infty.
\label{eq:lip}
\end{equation}

Let $\mathcal N_r$ be an $\ell_\infty$ $r$--net of $[0,1]^d$  \cite{vershynin2018high},
then $|\mathcal N_r|\le(3/r)^d$ Lemma 5.2 \cite{vershynin2018high}.
Using the $G$--Lipschitzness of $f_t,g_t$, from Lemma 5.2 \cite{vershynin2018high}
\begin{equation}
\sup_x\|\hat\mu_T(x)-\mu(x)\|_\infty
\le \max_{x\in\mathcal N_r}\|\hat\mu_T(x)-\mu(x)\|_\infty + 2G\,r.
\label{eq:net}
\end{equation}
For each fixed $x\in\mathcal N_r$ and coordinate $j\in\{1,2\}$,
Hoeffding’s inequality for $[0,B]$--bounded variables gives
\[
\Pr\!\left(\big|\hat\mu_T^{(j)}(x)-\mu^{(j)}(x)\big|>\gamma\right)
\le
2\exp\!\left(-\frac{2T \gamma^2}{B^2}\right).
\]
A union bound over $\mathcal N_r$ and the two coordinates yields
\begin{equation}
\Pr\!\left(
\max_{x\in\mathcal N_r}\|\hat\mu_T(x)-\mu(x)\|_\infty> \gamma
\right)
\le 4|\mathcal N_r|
\exp\!\left(-\frac{2T \gamma^2}{B^2}\right).
\label{eq:tail}
\end{equation}

Integrating the tail probability in \eqref{eq:tail} gives
\begin{align}
\mathbb E\!\left\{\max_{x\in\mathcal N_r}
\|\hat\mu_T(x)-\mu(x)\|_\infty\right\}
&\le
B\sqrt{\frac{\log(4|\mathcal N_r|)}{2T}}
+\frac{\sqrt{\pi}\,B}{2\sqrt{2T}}.
\label{eq:expnet}
\end{align}
Combining \eqref{eq:net} and \eqref{eq:expnet} yields
\begin{equation}
\mathbb E\!\left\{\sup_x\|\hat\mu_T(x)-\mu(x)\|_\infty\right\}
\le
B\sqrt{\frac{\log(4|\mathcal N_r|)}{2T}}
+\frac{\sqrt{\pi}\,B}{2\sqrt{2T}}
+2G\,r.
\label{eq:expall}
\end{equation}

Take $r=T^{-1/2}$, so $|\mathcal N_r|\le(3\sqrt{T})^d$ and
$\log(4|\mathcal N_r|)\le c_0 d + \tfrac{d}{2}\log T$
with $c_0=\log 12$.
Then \eqref{eq:expall} gives
\[
\mathbb E\!\left\{\sup_x\|\hat\mu_T(x)-\mu(x)\|_\infty\right\}
\le
C_1\,B\sqrt{\frac{d\log T}{T}}
+C_2 \frac{1}{\sqrt{T}}+ C_3\,\frac{G}{\sqrt{T}},
\]
for constants $C_1,C_2, C_3$.

Taking expectations in \eqref{eq:lip} and multiplying by $T$,
\begin{equation}\label{eq:finalbenchmarkdiff}
\mathbb{E}\left\{\big|C_\opt\}-T\Phi\big|\right\}
\le
T\,
\mathbb E\!\left\{\sup_x\|\hat\mu_T(x)-\mu(x)\|_\infty\right\}
\le
C_1\,B\,\sqrt{d\,T\,\log T}
+C_2\sqrt{T}+C_3\,G\,\sqrt{T}.
\end{equation}

\begin{remark} If there are $K$ functions $f_t^k$, $k=1,\dots,K$, instead of just two $f_t, g_t$, the bound  \eqref{eq:finalbenchmarkdiff} changes by an additive term $O(\sqrt{T \log K})$, resulting in the final bound of $O(\sqrt{T \log (TK)})$.
\end{remark}

\section{Proof of Lemma \ref{lem:R1}}\label{proofLemmaR1}
Let \(S_1:=\sum_{t=1}^T f_t(x_t)\) and \(S_2:=\sum_{t=1}^T g_t(x_t)\), and recall that $\lambda_t = (f_t(x_t), g_t(x_t))$ and \(\sum_t \Lambda_t(x_t,\theta_t)= \theta_{t1} f_t + \theta_{t2}g_t  =\sum_t\langle\theta_t,\lambda_t\rangle\). Then $$\max_{\theta}\sum_t \Lambda_t(x_t,\theta)=\max\{S_1,S_2\}.$$ 
Standard {\it Hedge} (gains version) \cite{cesa2006prediction} analysis implies that for any expert \(i=1,2\).
\[
\sum_{t=1}^T\langle\theta_t,\lambda_t\rangle \ge S_i - \frac{\ln 2}{\eta_{\theta,T}} - \sum_{t=1}^T \frac{\eta_{\theta,t}}{8} (\lambda_{t,1}-\lambda_{t,2})^2.
\]
Maximizing over \(i\) and using \((\lambda_{t,1}-\lambda_{t,2})^2\le (2B)^2=4B^2\) (since values are bounded by \(B\) from Assumption \ref{cvx}) yields
\[
\sum_{t=1}^T\langle\theta_t,\lambda_t\rangle \ge \max\{S_1,S_2\} - \frac{\ln 2}{\eta_{\theta,T}} - \sum_{t=1}^T\frac{\eta_{\theta,t} 4B^2 }{8}.
\]
Rearranging gives
\[
R_1= \max_{\theta\in\Delta_2}\sum_{t=1}^T \Lambda_t(x_t,\theta)-\sum_{t=1}^T \Lambda_t(x_t,\theta_t) =  \max\{S_1,S_2\} - \sum_{t=1}^T\langle\theta_t,\lambda_t\rangle \le \frac{\ln 2}{\eta_{\theta,T}} + \sum_{t=1}^T\frac{\eta_{\theta,t} B^2 }{2}.
\]

\section{Proof of Lemma \ref{lem:R2}}\label{proofLemmaR2}
Define \(\tilde f_t(x):=\Lambda_t(x,\theta_t)\), which is convex for fixed $\theta_t$. Thus, the standard OGD (projected) guarantee from \cite{HazanBook} implies, for any \(x\in\mathcal X\),
\[
\sum_{t=1}^T\tilde f_t(x_t)-\sum_{t=1}^T\tilde f_t(x)
\le \frac{\|x-x_0\|^2}{2\eta_{x,T}} + \sum_{t=1}^T \frac{\eta_{x,t}}{2} \|\nabla  {\tilde f}_t(x_{t-1})\|^2.
\]
Using the finite diameter \(\|x-x_0\|\le D\) and the bounded gradient \(\|\nabla  {\tilde f}_t(x_{t-1})\|\le G\) (since both 
 \(\|\nabla  f_t(x_{t-1})\|\le G\) and  \(\|\nabla  g_t(x_{t-1})\|\le G\)  from Assumption \ref{cvx}) yields
\[
R_2 \le \frac{D^2}{2\eta_{x,T}} + \sum_{t=1}^T \frac{\eta_{x,t} G^2 }{2}.
\]
With the choice \(\eta_{x,t} = D/(G\sqrt{t})\) 
\[
R_2 \le DG\sqrt{T}.
\]

\section{Proof of Lemma \ref{lem:R3}}\label{proofLemmaR3}

Recall that 
\[
\lambda_t(x) := (f_t(x), g_t(x)), \qquad \text{and let} \quad \ \mu(x) := \mathbb{E}\{\lambda_t(x)\}.
\]
where $\mu(x)$ is independent of index $t$ on account of the i.i.d. assumption on functions $f_t,g_t$.
Let $\mathcal{F}_{t-1}$ denote the filtration generated by
\[
\mathcal{F}_{t-1} := \sigma\big( x_1,\dots,x_{t-1}, \theta_1,\dots,\theta_{t-1}, \lambda_1(x_1),\dots,\lambda_{t-1}(x_{t-1}) \big),
\]
which determines $\theta_t$ at round $t$.

Recall the definition of optimal pair $x^\star$ and $\theta^\star$ that defines the static offline optimal solution of the expected {\it min-max} problem \eqref{defn:expregretintro}:
\begin{equation}\label{eq:dummyR31}
x^\star := \arg \min_{x \in \mathcal{X}} \max_{\theta \in \Delta_2} \sum_{t=1}^T \theta^\top \mu(x), \quad \text{and} \quad \theta^\star := \arg \max_{\theta \in \Delta_2} \sum_{t=1}^T \theta^\top \mu(x^\star).
\end{equation}

By the definition of $R_3$, 
\begin{align}\nn
R_3 
&= \min_{x\in \cX} \sum_{t=1}^T \theta_t^\top \lambda_t(x) 
- T \min_{x\in\cX} \max_{\theta \in \Delta_2}  \theta^\top \mu(x)  \\\nn
&= \min_{x\in \cX} \sum_{t=1}^T \theta_t^\top \lambda_t(x) 
- T \, \theta^\star{}^\top \mu(x^\star), \\ \label{eq:dummyR32}
&\stackrel{(b)}\le \sum_{t=1}^T \theta_t^\top \lambda_t(x^\star) 
- T \, \theta^\star{}^\top \mu(x^\star), 
\end{align}
where $(b)$ follows since $x^\star \in \cX$ is feasible.

Now rewrite the right-hand side of \eqref{eq:dummyR32} as
\begin{align}
\sum_{t=1}^T \theta_t^\top \lambda_t(x^\star) 
- T \, \theta^\star{}^\top \mu(x^\star)
&=
\sum_{t=1}^T (\theta_t - \theta^\star)^\top \lambda_t(x^\star)
\;+\;
\sum_{t=1}^T \theta^\star{}^\top \big( \lambda_t(x^\star) - \mu(x^\star) \big).
\label{eq:dummyR32b}
\end{align}

Moreover, we have 
\begin{equation}\label{eq:dummyR33}
\mathbb{E}\left\{(\theta_t - \theta^\star)^\top \lambda_t(x^\star) \,\Big|\, \mathcal{F}_{t-1} \right\} 
= (\theta_t - \theta^\star)^\top \bbE\{ \lambda_t(x^\star) \mid \mathcal{F}_{t-1}\} 
= (\theta_t - \theta^\star)^\top \mu(x^\star),
\end{equation}
where the first equality follows since $\theta_t$ is completely determined by $\mathcal{F}_{t-1}$, while the second equality follows from the i.i.d.\ assumption.

We also have 
\[
(\theta_t - \theta^\star)^\top \mu(x^\star) \le 0
\]
for each $t = 1, \dots, T$, by the optimality of $\theta^\star$ in
\[
\theta^\star \in \arg\max_{\theta \in \Delta_2} \theta^\top \mu(x^\star).
\]

Further, note that
\[
\bbE\!\left\{ \theta^\star{}^\top \big( \lambda_t(x^\star) - \mu(x^\star) \big) \,\Big|\, \mathcal{F}_{t-1} \right\} = 0,
\]
again by the i.i.d.\ assumption.

Combining the above and taking conditional expectations in \eqref{eq:dummyR32b}, we obtain
\begin{equation}\label{eq:dummyR34}
\bbE\!\left\{
\sum_{t=1}^T \theta_t^\top \lambda_t(x^\star) 
- T \, \theta^\star{}^\top \mu(x^\star)
\right\}
\le 0.
\end{equation}

Hence,
\begin{equation}
\mathbb{E}\{R_3\} \le 0.
\end{equation}

\section{Per-slot ${\it {\it min-max}}$ multi-objective optimization  \cite{lee2022online}}\label{app:benchmarkgap}

In this section, we show that benchmark $W_L^T $ \eqref{benchmarkdynamicminmax}  can be 
$3/2$ times more than our benchmark $C_\opt$ \eqref{defn:optcostintro}. 


\paragraph{Input}
Let $T=2N$ and $\mathcal{X}=[0,1]$. For $j=1,\dots,N$ let
\[
f_{2j-1}(x)=1.2-0.2x,\quad f_{2j}(x)=x,\qquad
g_{2j-1}(x)=x,\quad g_{2j}(x)=0.8+0.2x.
\]
Odd time slots: \(\min_{x\in[0,1]}\max\{1.2-0.2x,x\}=1\) (attained at \(x=1\)).  
Even time slots rounds: \(\min_{x\in[0,1]}\max\{x,0.8+0.2x\}=0.8\) (attained at \(x=0\)).  
Thus the benchmark $W_L^T $\eqref{benchmarkdynamicminmax}  is
\[
\sum_{t=1}^T \min_{x\in\mathcal{X}}\max\{f_t(x),g_t(x)\}=N(1+0.8)=1.8N.
\]
For any fixed $x$, for a pair of time slots $2j-1, 2j$, the sequence $f_t$ and $g_t$'s cost is 
\[
\sum_{t=2j-1}^{2j}f_t(x)=1.2+0.8x,\qquad
\sum_{t=2j-1}^{2j}g_t(x)=0.8+1.2x.
\]
Thus, $C_\opt$ \eqref{defn:optcostintro} is 
\[
C_\opt = \min_{x\in\mathcal{X}}\max\left\{\sum_{j=1}^{N}\sum_{t=2j-1}^{2j}1.2+0.8x, \sum_{j=1}^{N}\sum_{t=2j-1}^{2j} 0.8+1.2x\right\}=1.2N\quad(\text{at }x^\star=0).
\]
Hence, 
\[
\frac{W_L^T}{C_\opt}=\frac{1.8N}{1.2N}=1.5,
\]
as claimed.
 Therefore, even if an online algorithm $\cA$ achieves sub-linear regret with respect to the benchmark $W_L^T $\eqref{benchmarkdynamicminmax}, it does not necessarily have 
 sub-linear regret for our benchmark $C_\opt$ \eqref{defn:optcostintro}.
\begin{remark}\label{rem:sisalem}
{\bf Detailed comparison with $\alpha$-fair model of \cite{si2023enabling}} The transition from $\alpha$-fairness (with finite $\alpha$) \eqref{defn:U} to the {\it min-max} objective \eqref{defn:costAintro} marks a shift from a smooth, strictly concave optimization landscape to a non-smooth, polyhedral one. For any finite $\alpha \ge 0$, the $\alpha$-fair utility function $U_{\alpha}(z)$ \eqref{defn:U} acts as a differentiable, strictly concave regularizer that ``softens" the trade-offs between competing objectives. In the finite $\alpha$ regime, the gradient of the aggregate utility changes continuously with the input, allowing algorithms to rely on the local curvature of the utility function to guide the learning process. However, as $\alpha \to \infty$, the utility function converges to the {\it min-max} objective, which is equivalent to an $L_{\infty}$ norm. Geometrically, the smooth, rounded contours of the $\alpha$-fair utility collapse into a sharp wedge where the derivative is discontinuous at the points where multiple objectives have equal costs. This loss of differentiability explains why prior regret bounds of $O(\alpha\sqrt{T})$ \cite{si2023enabling} become vacuous in the {\it min-max} limit; they are inversely proportional to the curvature of the utility, which vanishes as the objective becomes piecewise linear. 

In comparison, the {\it Hedge+OGD} algorithmic approach succeeds as $\alpha\rightarrow \infty$ because it does not attempt to ``smoothen" this non-differentiable cusp. Instead, by employing the {\it Hedge} algorithm to update weights $\theta_t$, it treats the selection of the dominant objective as a discrete experts problem. This effectively bypasses the numerical instability inherent in the $\alpha \to \infty$ transition and provides a regret guarantee that remains independent of the parameter $\alpha$.

{\it Hedge+OGD} algorithm is also applicable for finite $\alpha$, however, since it is designed to remain stable in the non-smooth limit ($\alpha \to \infty$), it lacks the specialized mechanics required to exploit the curvature present when $\alpha$ is small. 
For finite $\alpha$, the $\alpha$-fair utility function is strictly concave, offering a smooth landscape for optimization. The OHF algorithm \cite{si2023enabling} is specifically tailored to exploit this strict concavity, allowing the algorithm to achieve a regret bound of $O(\alpha\sqrt{T})$. When $\alpha$ is small, the utility function possesses significant curvature, which provides a strong, directional gradient signal that accelerates convergence. {\it Hedge} on the other hand is a first-order expert algorithm designed for the simplex; it does not naturally incorporate the second-order information or the strong concavity of the underlying utility. Consequently, it cannot ``speed up" when the landscape is smooth. 

The main takeaway is that 
the finite $\alpha$ and $\alpha \rightarrow \infty$ regimes are fundamentally different and require different treatment.
\end{remark}

\section{Proof of Theorem~\ref{thm:bandit}}\label{app:banditproof}
Recall the decomposition~\eqref{defn:regretdecomp},
\[
{\bar R}_\cA(T)
=
\sup_{\cD}\{\bbE\{R_1\} + \bbE\{R_2\} + \bbE\{R_3\}\}
\;\le\;
\sup_{\cD}\{\bbE\{R_1\}\}
+
\sup_{\cD}\{\bbE\{R_2\}\}
+
\sup_{\cD}\{\bbE\{R_3\}\}.
\]

\paragraph{Bounding $R_1$ and $R_3$.}
Recall that
\[
R_1
:=
\max_{\theta\in\Delta_K}
\sum_{t=1}^T \Lambda_t(x_t,\theta)
-
\sum_{t=1}^T \Lambda_t(x_t,\theta_t),
\qquad
\Lambda_t(x,\theta)=\sum_{k=1}^K \theta_k f_t^k(x).
\]

In the bandit setting, the Hedge update observes the queried losses
$\{f_{t-1}^k(\tilde x_t)\}_{k=1}^K$.
To relate $R_1$ to the Hedge regret on the queried points, we proceed as follows.

\medskip
\noindent
\emph{Step 1 (add and subtract queried losses inside the maximization).}
For any fixed $\theta\in\Delta_K$,
\[
\sum_{t=1}^T \Lambda_t(x_t,\theta)
=
\sum_{t=1}^T \Lambda_t(\tilde x_t,\theta)
+
\sum_{t=1}^T\bigl(
\Lambda_t(x_t,\theta)-\Lambda_t(\tilde x_t,\theta)
\bigr).
\]
Substituting this identity into the definition of $R_1$ yields
\[
\begin{aligned}
R_1
&=
\max_{\theta\in\Delta_K}
\Bigg[
\sum_{t=1}^T \Lambda_t(\tilde x_t,\theta)
-
\sum_{t=1}^T \Lambda_t(x_t,\theta_t)
\\
&\hspace{3cm}
+
\sum_{t=1}^T\bigl(
\Lambda_t(x_t,\theta)-\Lambda_t(\tilde x_t,\theta)
\bigr)
\Bigg].
\end{aligned}
\]

\medskip
\noindent
\emph{Step 2 (split the maximization).}
Using the elementary inequality
\[
\max_{\theta}(a_\theta+b_\theta)
\;\le\;
\max_{\theta} a_\theta + \max_{\theta} b_\theta,
\]
we obtain
\[
\begin{aligned}
R_1
&\le
\max_{\theta\in\Delta_K}
\sum_{t=1}^T \Lambda_t(\tilde x_t,\theta)
-
\sum_{t=1}^T \Lambda_t(x_t,\theta_t)
\\
&\quad
+
\max_{\theta\in\Delta_K}
\sum_{t=1}^T\bigl(
\Lambda_t(x_t,\theta)-\Lambda_t(\tilde x_t,\theta)
\bigr).
\end{aligned}
\]

\medskip
\noindent
\emph{Step 3 (Lipschitz bound, uniform in $\theta$).}
By Assumption~3, each $f_t^k$ is $G$-Lipschitz, and since
$\theta\in\Delta_K$ is a convex combination,
\[
\bigl|
\Lambda_t(x_t,\theta)-\Lambda_t(\tilde x_t,\theta)
\bigr|
\le
G\|x_t-\tilde x_t\|,
\qquad
\forall\,\theta\in\Delta_K.
\]
Therefore,
\[
\max_{\theta\in\Delta_K}
\sum_{t=1}^T\bigl(
\Lambda_t(x_t,\theta)-\Lambda_t(\tilde x_t,\theta)
\bigr)
\le
G\sum_{t=1}^T \|x_t-\tilde x_t\|.
\]

Combining the above bounds gives
\[
R_1
\le
\max_{\theta\in\Delta_K}
\sum_{t=1}^T \Lambda_t(\tilde x_t,\theta)
-
\sum_{t=1}^T \Lambda_t(x_t,\theta_t)
+
G\sum_{t=1}^T \|x_t-\tilde x_t\|.
\]

\medskip
\noindent
\emph{Step 4 (remove the remaining mismatch outside the maximization).}
Adding and subtracting
$\sum_{t=1}^T \Lambda_t(\tilde x_t,\theta_t)$ yields
\[
\begin{aligned}
R_1
&\le
\Bigg[
\max_{\theta\in\Delta_K}
\sum_{t=1}^T \Lambda_t(\tilde x_t,\theta)
-
\sum_{t=1}^T \Lambda_t(\tilde x_t,\theta_t)
\Bigg]
\\
&\quad
+
\Bigg[
\sum_{t=1}^T \Lambda_t(\tilde x_t,\theta_t)
-
\sum_{t=1}^T \Lambda_t(x_t,\theta_t)
\Bigg]
+
G\sum_{t=1}^T \|x_t-\tilde x_t\|.
\end{aligned}
\]

The second bracket is again bounded by Lipschitz continuity:
\[
\bigl|
\Lambda_t(\tilde x_t,\theta_t)-\Lambda_t(x_t,\theta_t)
\bigr|
\le
G\|x_t-\tilde x_t\|.
\]

\medskip
\noindent
\emph{Final bound on $R_1$.}
\[
R_1
\le
\underbrace{
\max_{\theta\in\Delta_K}
\sum_{t=1}^T \Lambda_t(\tilde x_t,\theta)
-
\sum_{t=1}^T \Lambda_t(\tilde x_t,\theta_t)
}_{\text{Hedge regret on queried losses}}
\;+\;
2G\sum_{t=1}^T \|x_t-\tilde x_t\|.
\]

The first term is exactly the regret of the $K$-experts problem with gains
$f_{t-1}^k(\tilde x_t)$, and hence with
$\eta_{\theta,t}=\sqrt{\log K}/(B\sqrt{t})$,
\[
\max_{\theta\in\Delta_K}
\sum_{t=1}^T \Lambda_t(\tilde x_t,\theta)
-
\sum_{t=1}^T \Lambda_t(\tilde x_t,\theta_t)
\le
B\sqrt{2T\log K}.
\]

Moreover, $\|x_t-\tilde x_t\|\le \delta$, so the additional term
$2G\sum_t\|x_t-\tilde x_t\|$ is of order $\sum_t\delta$. We upper bound this in the two cases as follows.

\begin{itemize}
\item
\textbf{One-point feedback.}
With $\delta = T^{-1/4}$,
\[
2G\sum_{t=1}^T \delta
=
O\!\left(G\sum_{t=1}^T T^{-1/4}\right)
=
O(GT^{3/4}).
\]

\item
\textbf{Two-point feedback.}
With $\delta = T^{-1/2}$,
\[
2G\sum_{t=1}^T \delta
=
O\!\left(G\sum_{t=1}^T T^{-1/2}\right)
=
O(G\sqrt{T}).
\]
\end{itemize}

Note that the term $R_3$ depends only on the true loss functions $(f_t,g_t)$ and the sequence
$\{\theta_t\}_{t=1}^T$, and not on how gradients are estimated; hence the proof of
$\bbE\{R_3\} \le 0$ from Appendix~\ref{proofLemmaR3} applies verbatim in the bandit setting as well.

Thus, \[
\bbE\{R_3\} \le 0.
\]

We bound  $R_2$ as follows.

{\bf Bounding $R_2$ (one-point feedback).}
Using $\delta = T^{-1/4}$ and $\eta_{x,t}= D/(G d^{1/2} t^{3/4})$, $\rho=\delta/\sfr$ where $r\cB_1\subseteq \cX \subseteq D\cB_1$ and $\cB_1$ is the unit ball,   
the bandit-OGD analysis \cite{flaxman2005bandit} yields
\[
R_2 \leq c_1 \, d^{1/2} G D \, T^{3/4},
\]
for some universal constant $c_1$.

Combining the three bounds we get 
\[
{\bar R}_\cA(T) \leq c_1 \, d^{1/2} G D \, T^{3/4} \;+\; 2B\sqrt{2T\log K}.
\]

{\bf Bounding $R_2$ (two-point feedback).}
Choosing $\delta = T^{-1/2}$, $\rho=\delta/\sfr$ and $\eta_{x,t}= D/(G d^{1/2}\sqrt{t})$, 
the two-point bandit-OGD analysis \cite{agarwal2010bandit, duchi2015optimal, shamir2017optimal} gives
\[
R_2 \leq c_3 \, d^{1/2} G D \, \sqrt{T},
\]
for some universal constant $c_3$.  Thus, with two-point feedback,
\[
{\bar R}_\cA(T) \leq c_3 \, d^{1/2} G D \, \sqrt{T} \;+\; 2B\sqrt{2T\log K}.
\]

\section{Experts Problem (Gains Version)}\label{sec:hedgeintro}

We are given $N$ experts. The interaction proceeds for $T$ rounds as follows:

\begin{itemize}
    \item On each round $t = 1,2,\dots,T$:
    \begin{enumerate}
        \item Each expert $i \in \{1,\dots,N\}$ receives a gain $g_{t,i} \in [0,1]$.
        \item The learner chooses a probability distribution 
        \[
            p_t = (p_{t,1}, \dots, p_{t,N}),
        \]
        where $p_{t,i} \geq 0$ and $\sum_{i=1}^N p_{t,i} = 1$.
        \item The learner’s gain is the expected value
        \[
            G_t = \sum_{i=1}^N p_{t,i} \, g_{t,i}.
        \]
    \end{enumerate}
\end{itemize}

\noindent
The learner’s goal is to compete with the best fixed expert in hindsight by minimizing the regret
\[
R_T = \max_{i} \sum_{t=1}^T g_{t,i} \;-\; \sum_{t=1}^T G_t .
\]

\section*{Hedge Algorithm (Gains Version)}

\begin{algorithm}[H]
\caption{Hedge (Gains Version)}
\begin{algorithmic}[1]
\Require Number of experts $N$, learning rate $\eta > 0$
\State Initialize weights: $w_{1,i} \gets 1$ for all $i \in \{1,\dots,N\}$
\For{$t = 1,2,\dots,T$}
    \State Form probabilities:
    \[
        p_{t,i} \gets \frac{w_{t,i}}{\sum_{j=1}^N w_{t,j}} \quad \forall i
    \]
    \State Play according to distribution $p_t = (p_{t,1},\dots,p_{t,N})$
    \State Observe gains $g_{t,i} \in [0,1]$ for each expert $i$
    \State Update weights:
    \[
        w_{t+1,i} \gets w_{t,i} \, e^{\eta g_{t,i}} \quad \forall i
    \]
\EndFor
\end{algorithmic}
\end{algorithm}

\section*{Guarantee}
\begin{theorem} \cite{cesa2006prediction}
With $\eta = \sqrt{\tfrac{2 \ln N}{T}}$, Hedge achieves
\[
\max_{i} \sum_{t=1}^T g_{t,i} - \sum_{t=1}^T G_t \;\;\leq\;\; \sqrt{2 T \ln N}.
\]
\end{theorem}

\section{Proof of Theorem \ref{thm:lbbinary}}\label{app:lbbinary}
Given the context is clear in this Appendix, we drop superscript $b$ from both cost $C^b_\cA$ and $C^b_\opt$ and write it simply as $C_\cA$ and $C_\opt$, respectively.
\begin{proof} Recall that at each round $t=1,\dots,T$, the algorithm $\mathcal A$ predicts
$x_t\in\{0,1\}$ and observes bits $(b_t^1,b_t^2)\in\{0,1\}^2$.
The losses are
\[
\ell_t^1(x_t) = \mathbf 1\{x_t\neq b_t^1\},
\qquad
\ell_t^2(x_t) = \mathbf 1\{x_t\neq b_t^2\}.
\]

The algorithm’s total cost on an input sequence $\sigma$ is
\[
C_{\mathcal A}(\sigma)
=
\max\Bigl\{
\sum_{t=1}^T \ell_t^1(x_t),
\;
\sum_{t=1}^T \ell_t^2(x_t)
\Bigr\}.
\]

The static benchmark is
\[
C_\opt(\sigma)
=
\min_{x\in\{0,1\}}
\max\Bigl\{
\sum_{t=1}^T \mathbf 1\{x\neq b_t^1\},
\;
\sum_{t=1}^T \mathbf 1\{x\neq b_t^2\}
\Bigr\}.
\]

\subsection{Adversarial construction}

Assume $T$ is even.
The adversary constructs two inputs $\sigma_1$ and $\sigma_2$ that are
identical for the first $T/3$ rounds.

\paragraph{Common prefix.}
For $t=1,\dots,T/3$,
\[
b_t^1 = 0,
\qquad
b_t^2 = 1.
\]

Let
\[
\bar x := \sum_{t=1}^{T/3} x_t
\]
denote the number of times the algorithm predicts $1$ during the prefix.

\paragraph{Suffix selection rule.}
After observing $\bar x$, the adversary completes the input as follows:
\begin{itemize}
\item If $\bar x > T/6$, the adversary chooses input $\sigma_1$.
\item If $\bar x \le T/6$, the adversary chooses input $\sigma_2$.
\end{itemize}

\paragraph{Suffix of $\sigma_1$.}
For $t=T/3+1,\dots,T$:
\[
b_t^1 \sim \mathrm{Bernoulli}\Bigl(\tfrac12\Bigr) \;\text{i.i.d.},
\qquad
b_t^2 = 0.
\]

\paragraph{Suffix of $\sigma_2$.}
For $t=T/3+1,\dots,T$:
\[
b_t^1 = 1,
\qquad
b_t^2 \sim \mathrm{Bernoulli}\Bigl(\tfrac12\Bigr) \;\text{i.i.d.}.
\]

\subsection{Cost of the static benchmark}

We compute $C_\opt(\sigma_i)$ for $i=1,2$.
With $\sigma_1$, $\opt$ predicts $x=0$ always, and 
$$\bbE\{C_\opt(\sigma_1)\} = 
\bbE\left\{\max\left\{\sum_{t=1}^{T/3} \mathbf 1\{0\neq b_t^1\} + \sum_{t=T/3+1}^{T} \mathbf 1\{0\neq b_t^1\}, \ \ \sum_{t=1}^{T} \mathbf 1\{0\neq b_t^2\}\right\}\right\} 
\le  T/3 +o(T).$$

Similarly, with $\sigma_2$, $\opt$ predicts $x=1$ always, and 
$$\bbE\{C_\opt(\sigma_2)\} = 
\bbE\left\{\max\left\{ \sum_{t=1}^{T} \mathbf 1\{1\neq b_t^1\}, \ \  \sum_{t=1}^{T/3} \mathbf 1\{1\neq b_t^2\} + \sum_{t=T/3+1}^{T} \mathbf 1\{1\neq b_t^2\}\right\}\right\} 
\le  T/3 +o(T).$$

\subsection{Lower bound on any online algorithm’s cost}

We now lower bound the expected cost of $\mathcal A$.

\paragraph{Case 1: $\bar x > T/6$ (input $\sigma_1$).}
For the common prefix, 
\[
\sum_{t\le T/3} \ell_t^1(x_t) = \bar x,
\qquad
\sum_{t\le T/3} \ell_t^2(x_t) = \frac{T}{3}-\bar x.
\]

In the suffix of $\sigma_1$, $b_t^1$ is unbiased and independent of $x_t$, so
\[
\mathbb E\!\left\{\sum_{t>T/3} \ell_t^1(x_t)\right\} = \frac{T}{3}.
\]

Thus,
\begin{equation}\label{eq:binST1}
\mathbb E\!\left\{\sum_{t=1}^T \ell_t^1(x_t)\right\} = \bar x+ \frac{T}{3}.
\end{equation}

Also,
\[
\sum_{t>T/3} \ell_t^2(x_t) = \sum_{t>T/3} x_t \ge 0.
\]

By definition, 
$$
\bbE\{C_{\mathcal A}(\sigma_1)\}
\;\ge\; \mathbb E\!\left\{\sum_{t=1}^T \ell_t^1(x_t)\right\}.$$ Thus, using \eqref{eq:binST1} and the condition that $\bar x>T/6$, we have 
$$
\bbE\{C_{\mathcal A}(\sigma_1)\}
\;\ge\  {\bar x}+ \frac{T}{3}=\frac{T}{2}.$$

\paragraph{Case 2: $\bar x \le T/6$ (input $\sigma_2$).}
By symmetry,
\[
\bbE\{C_{\mathcal A}(\sigma_2)\}
\ge \mathbb E\!\left\{\sum_{t=1}^T \ell_t^2(x_t)\right\} =  \Bigl(\frac{T}{3}-\bar x\Bigr) + \frac{T}{3} \ge 
 \Bigl(\frac{T}{3}-\frac{T}{6}\Bigr) + \frac{T}{3} = \frac{T}{3}+\frac{T}{6}  = \frac{T}{3}+ \frac{T}{6}
\]
since $\bar x\le T/6$.

Thus, in both cases
\[
\bbE\{C_{\mathcal A}(\sigma_i)\}
\ge
\frac{T}{3} + \Omega(T),
\]
which together with $\bbE\{C_\opt(\sigma_1)\}=\bbE\{C_\opt(\sigma_2)\}\le \frac{T}{3} +o(T)$ imply that the static {\it min-max} regret $R^b_\cA(T)$ for any online algorithm is $\Omega(T)$.

\end{proof}

\section{Deriving $\bbE\{R^b_\cA(T)\}=O(\sqrt{T\log T})$ for the binary prediction problem using Algorithm \ref{alg:main}}\label{app:binaryspecialcase}
Let an online algorithm $\cA$ choose $x_t=1$ with probability $p_t\in[0,1]$ and
$x_t=0$ with probability $1-p_t$.  Define the expected losses
\[
f_t(p)
:=\mathbb E[\mathbf 1\{x_t\neq f_t\}]
=
|p-f_t|,
\qquad
g_t(p)
:=\mathbb E[\mathbf 1\{x_t\neq g_t\}]
=
|p-g_t|.
\]
Both $f_t(p)$ and $g_t(p)$ are convex, 1-Lipschitz functions of $p$, thus, 
exactly matching the setup of Section~\ref{sec:iid} with convex losses. Thus, if we use Algorithm \ref{alg:main} to find probability $p_t$ for choosing $x_t=1$, 
Theorem \ref{thm:main} implies that the {\it min-max} static regret with the true benchmark $\bbE\{C_\opt\}$ is 
$\bbE\{C_\cA\}  - \bbE\{C_\opt\} \le  O(\sqrt{T \log T}).$
\section{Proof of Theorem \ref{thm:binary}}\label{app:dm}

We consider the  \textbf{Adaptive Stronger-Bias Double Majority (ASB-DM)} algorithm (Algorithm \ref{alg:binary}) to solve the {\it min-max} prediction of binary sequences.
The ASB-DM algorithm employs a hierarchical voting structure to resolve the online {\it min-max} objective. At each time step $t$, the algorithm maintains empirical means $\hat{q}_{1,t}$ and $\hat{q}_{2,t}$ for the two cost sequences. The decision rule is governed by two distinct layers of majority-based logic:

\begin{enumerate}
    \item \textbf{The First Majority: Marginal Consensus.} The algorithm first evaluates the marginal preference of each sequence. If both sequences independently reach a consensus—meaning both $\hat{q}_{1,t}, \hat{q}_{2,t} < 1/2$ (favoring action $0$) or both $\hat{q}_{1,t}, \hat{q}_{2,t} > 1/2$ (favoring action $1$)—the algorithm follows this local majority. In such cases, the optimal action is unambiguous as both sequences indicate the same superior choice.
    
    \item \textbf{The Second Majority: Stronger-Bias Resolution.} A fundamental challenge arises during \textit{disagreement scenarios}, where the sequences provide conflicting marginal majorities. To achieve a \textbf{double majority} that resolves this conflict, the algorithm evaluates the sum of empirical means, $\Sigma_t = \hat{q}_{1,t} + \hat{q}_{2,t}$. This sum identifies which of the two conflicting marginal preferences is statistically ``stronger.'' The threshold $\Sigma_t = 1$ acts as the secondary decision boundary: if $\Sigma_t < 1$, the magnitude of the evidence favors action $0$; if $\Sigma_t \ge 1$, the bias favors action $1$.
\end{enumerate}

The efficacy of this approach relies on the concentration of the empirical vector $(\hat{q}_{1,t}, \hat{q}_{2,t})$ toward the true parameters $(q_1, q_2)$. As long as the true parameters do not lie exactly on the boundary $q_1 + q_2 = 1$, the probability of the algorithm's double-majority resolution failing decays exponentially. This ensures that the decision process \textbf{stabilizes} after a finite number of steps, effectively bounding the total expected mistakes by a constant and yielding the $O(1)$ regret guarantee.

\begin{algorithm}[H]
\caption{Adaptive Stronger-Bias Double Majority (ASB-DM)}
\label{alg:binary}
\begin{algorithmic}[1]
\For{$t = 1, 2, \dots, T$}

    Compute empirical means $\hat{q}_{1,t}$ and $\hat{q}_{2,t}$.\;
    
    \If{$\hat{q}_{1,t} < 1/2$ \textbf{and} $\hat{q}_{2,t} < 1/2$}
    
        $x_t \gets 0$ \Comment{Case A: Agreement on 0}
    
    \ElsIf{$\hat{q}_{1,t} \ge 1/2$ \textbf{and} $\hat{q}_{2,t} \ge 1/2$}
    
        $x_t \gets 1$ \Comment{Case B: Agreement on 1}
    
    \Else{}\Comment{Case C: Disagreement} 
    
        Compute $\Sigma_t = \hat{q}_{1,t} + \hat{q}_{2,t}$.\;
        \If{$\Sigma_t < 1$}
        $x_t \gets 0$ \Comment{Bias favors 0}
        \Else \ \ 
            $x_t \gets 1$ \Comment{Bias favors 1}
        \EndIf
    \EndIf
    
    Play $x_t$.\;
\EndFor
\end{algorithmic}
\end{algorithm}

\begin{proof}[Proof of Theorem \ref{thm:binary}]
Let the cumulative loss $L_k(x)$ for a fixed static action $x \in \{0,1\}$ on sequence $b_t^k, k=1,2$ be 
\[
L_k(x) = \sum_{t=1}^T \ell_{t,k}(x) = \sum_{t=1}^T \mathbf{1}\{x \neq b_t^k\}.
\] 
Let $L_k(x_{1:T})$ denote the cumulative loss incurred by an algorithm $\cA$ actions $x_t$ on sequence $b_t^k, k=1,2$:
\[
L_k(x_{1:T}) = \sum_{t=1}^T \ell_{t,k}(x_t) = \sum_{t=1}^T \mathbf{1}\{x_t \neq b_t^k\}.
\]
Let $k^*$ be the index of the sequence that maximizes this cumulative loss:
\[
k^* = \operatorname*{arg\,max}_{k \in \{1,2\}} L_k(x_{1:T}).
\]

For further exposition, we distinguish between the true path-dependent optimal action and the optimal static action based on expected losses.
\begin{enumerate}
    \item Let $x^\star$ be the random variable representing the best action in hindsight for the specific realization of losses:
    \[
    x^\star = \arg\min_{x \in \{0,1\}} \max_{k \in \{1,2\}} L_k(x).
    \]

The true regret is defined as the difference between the algorithm's worst-case loss and the benchmark's worst-case loss:
\[
R_{\mathcal{A}}^{b}(T) = L_{k^*}(x_{1:T}) - \max_{k \in \{1,2\}} L_k(x^\star).
\]
Since the maximum loss of the benchmark is necessarily greater than or equal to its loss on the specific sequence $k^*$ (i.e., $\max_k L_k(x^\star) \ge L_{k^*}(x^\star)$), subtracting the latter yields a valid upper bound on the regret:
\begin{equation}\label{eq:equivregretbinary}
R_{\mathcal{A}}^{b}(T) \le L_{k^*}(x_{1:T}) - L_{k^*}(x^\star).
\end{equation}

    \item Let $\bar{x}^\star$ be the fixed static action that minimizes the worst-case \textit{expected} cumulative loss:
  \begin{equation}\label{eq:xbardefn}
    \bar{x}^\star = \arg\min_{x \in \{0,1\}} \max_{k \in \{1,2\}} \bbE\{L_k(x)\} =  \arg\min_{x \in \{0,1\}} \max_{k \in \{1,2\}} T\mathbb{P}(b_t^k \neq x). 
\end{equation}
\end{enumerate}

\textbf{Regret Decomposition:}
Using \eqref{eq:equivregretbinary}, we decompose the regret relative to the fixed reference action $\bar{x}^\star$:
\begin{align}\nn
R^b_\cA(T) &\le  L_{k^*}(x_{1:T}) - L_{k^*}(x^\star) \\ \label{eq:regretdecompbinary}
&= \underbrace{\left( L_{k^*}(x_{1:T}) - L_{k^*}(\bar{x}^\star) \right)}_{\text{(A) Algorithm's Mistakes}} + \underbrace{\left( L_{k^*}(\bar{x}^\star) - L_{k^*}(x^\star) \right)}_{\text{(B) Estimation Cost}}.
\end{align}

\noindent\textbf{Bounding Term (B):}
We define the expected worst-case loss rate for any fixed action $x$ as:
\[
\nu(x) = \max_{k \in \{1,2\}} \mathbb{P}(b_t^k \neq x).
\]
Since we assume the asymmetric case ($q_1+q_2 \neq 1$), the minimizer $\bar{x}^\star$ \eqref{eq:xbardefn} is unique. Let $x_{\text{alt}}$ denote the alternative action ($x_{\text{alt}} \neq \bar{x}^\star$). There exists a strictly positive gap $\delta > 0$ defined as:
\[
\delta = \nu(x_{\text{alt}}) - \nu(\bar{x}^\star) > 0.
\]
This gap $\delta$ is strictly positive because $\delta = 0$ implies $\nu(0) = \nu(1)$, which can only occur if $q_k = 1/2$ or $q_1+q_2=1$. Since our theorem assumes $q_1, q_2 \neq 1/2$ and $q_1+q_2 \neq 1$, the costs are distinct, ensuring a unique minimizer. Note that $\delta$ depends only on the fixed distribution parameters $q_1, q_2$ and is independent of the horizon $T$.

We now bound Term (B), which represents the estimation cost:
\[
\text{(B)} = V(\bar{x}^\star) - \min_{x \in \{0,1\}} V(x),
\]
where $V(x) = \max_k L_k(x)$ is the realized worst-case cumulative loss. Term (B) is positive only if the alternative action becomes empirically optimal (i.e., $V(x_{\text{alt}}) < V(\bar{x}^\star)$, meaning $x^\star \neq \bar{x}^\star$). Since the difference in losses is at most $T$, we bound the expectation as:
\[
\bbE\{\text{(B)}\} \le T \cdot \mathbb{P}( V(x_{\text{alt}}) < V(\bar{x}^\star) ).
\]
The event $\{ V(x_{\text{alt}}) < V(\bar{x}^\star) \}$ implies a significant deviation from the expected rates. Specifically, if for all $k$ and $x$, the empirical losses satisfy $|L_k(x)/T - \bbE\{L_k(x)/T\}| < \delta/2$, then:
\[
\frac{V(x_{\text{alt}})}{T} = \max_k \frac{L_k(x_{\text{alt}})}{T} > \nu(x_{\text{alt}}) - \frac{\delta}{2} = \nu(\bar{x}^\star) + \frac{\delta}{2} > \max_k \frac{L_k(\bar{x}^\star)}{T} = \frac{V(\bar{x}^\star)}{T}.
\]
This contradicts $V(x_{\text{alt}}) < V(\bar{x}^\star)$. Therefore, for the ordering to flip, i.e. $\{ V(x_{\text{alt}}) < V(\bar{x}^\star) \}$, at least one empirical loss $L_k(x)$ must deviate from its mean by at least $\delta/2$, i.e. $|L_k(x)/T - \bbE\{L_k(x)/T\}| \ge  \delta/2$. Applying the Union Bound over the 4 quantities ($k \in \{1,2\}, x \in \{0,1\}$) and Hoeffding's inequality:
\[
\mathbb{P}( V(x_{\text{alt}}) < V(\bar{x}^\star) ) \le \sum_{x,k} \mathbb{P}\left( \left| \frac{L_k(x)}{T} - \mathbb{E}\left[\frac{L_k(x)}{T}\right] \right| \ge \frac{\delta}{2} \right) \le 4 \cdot 2e^{-2T(\delta/2)^2}.
\]
Substituting this back into the expectation:
\begin{equation}\label{eq:termB}
\bbE\{\text{(B)}\} \le 8T e^{-\delta^2 T / 2} = o(1).
\end{equation}

\noindent\textbf{Bounding Term (A):}
We now upper bound Term (A), which is defined as $L_{k^*}(x_{1:T}) - L_{k^*}(\bar{x}^\star)$.
Writing in terms of $\ell_{t,k}(x_t)$, we get
\[
\text{(A)} = \sum_{t=1}^T \left( \ell_{t,k^*}(x_t) - \ell_{t,k^*}(\bar{x}^\star) \right).
\]
Consider the term inside the sum:
\begin{itemize}
    \item If $x_t = \bar{x}^\star$, then $\ell_{t,k^*}(x_t) - \ell_{t,k^*}(\bar{x}^\star) = 0$.
    \item If $x_t \neq \bar{x}^\star$, the difference is at most $1$ (since losses are binary).
\end{itemize}
Thus, regardless of the index $k^*$, the term is bounded by the indicator of a mistake:
\[
\ell_{t,k^*}(x_t) - \ell_{t,k^*}(\bar{x}^\star) \le \mathbf{1}\{x_t \neq \bar{x}^\star\}.
\]
Summing over $t$:
\begin{equation}\label{eq:binaryregretbound}
\text{(A)} \le \sum_{t=1}^T \mathbf{1}\{x_t \neq \bar{x}^\star\}.
\end{equation}
This bounds Term (A) by the total number of mistakes against the fixed reference.
Next, to bound the expected regret, we analyze the probability of error $\mathbb{P}(x_t \neq \bar{x}^\star)$. Note that while $b_t^1$ and $b_t^2$ may be correlated at any instant $t$, the vectors $\mathbf{b}_t = (b_t^1, b_t^2)$ are i.i.d. across time $t$.

\paragraph{Characterization of Optimal Static Action $\bar{x}^\star$.}
We explicitly determine $\bar{x}^\star$ by minimizing the worst-case expected cumulative loss derived in the definition.
For $x=0$, the loss is $T \cdot \mathbb{P}(b_t^k=1) = T q_k$. For $x=1$, the loss is $T \cdot \mathbb{P}(b_t^k=0) = T(1-q_k)$.
The {\it min-max} optimal action minimizes the maximum of these expected cumulative losses:
\[
\bar{x}^\star = \arg\min_{x \in \{0,1\}} \max \left( \bbE\{L_1(x)\}, \bbE\{L_2(x)\} \right).
\]
\begin{itemize}
    \item \textbf{Case I (Agreement):} If $q_1 < 1/2$ and $q_2 < 1/2$, the max cost of $x=0$ is $T \cdot \max(q_1, q_2) < 0.5T$. The max cost of $x=1$ is $T \cdot \max(1-q_1, 1-q_2) > 0.5T$. Thus, $\bar{x}^\star = 0$. By symmetry, if both $>1/2$, $\bar{x}^\star=1$.
    \item \textbf{Case II (Disagreement):} Assume w.l.o.g. $q_1 < 1/2 < q_2$.
    The max cost of $x=0$ is determined by sequence 2 (since $q_2 > q_1$):
    \[ \max(T q_1, T q_2) = T q_2. \]
    The max cost of $x=1$ is determined by sequence 1 (since $1-q_1 > 1-q_2$):
    \[ \max(T(1-q_1), T(1-q_2)) = T(1-q_1). \]
    Therefore, $\bar{x}^\star = 0$ if and only if $T q_2 < T(1-q_1)$, which simplifies to $q_1 + q_2 < 1$.
\end{itemize}

\subparagraph{Case I: When either both $q_1 < 1/2$ and $q_2 < 1/2$ or $q_1 > 1/2$ and $q_2 > 1/2$.}
Assume w.l.o.g. that $q_1 < 1/2$ and $q_2 < 1/2$; the other case follows similarly. The optimal action is $\bar{x}^\star = 0$. The algorithm plays $x_t = 0$ if $\hat{q}_{1,t} < 1/2$ and $\hat{q}_{2,t} < 1/2$. Conversely, a mistake ($x_t = 1$) implies at least one empirical mean has deviated:
\[
\{x_t = 1\} \subseteq \{ \hat{q}_{1,t} \ge 1/2 \} \cup \{ \hat{q}_{2,t} \ge 1/2 \}.
\]
Let $\Delta = \min(1/2 - q_1, 1/2 - q_2) > 0$. We apply the union bound and Hoeffding's inequality to the marginal probabilities:
\[
\mathbb{P}(x_t \neq \bar{x}^\star) \le \mathbb{P}(\hat{q}_{1,t} \ge 1/2) + \mathbb{P}(\hat{q}_{2,t} \ge 1/2) \le 2e^{-2(t-1)\Delta^2}.
\]

\subparagraph{Case II: When either $q_1 < 1/2 < q_2$ or $q_2 < 1/2 < q_1$ while $q_1 + q_2 < 1$.}
Recall the definition of {\it disagreement}, denoted by $\mathcal{D}_t$, which holds if the empirical means imply conflicting actions:
\[
\mathcal{D}_t = 
\left\{ 
\hat{q}_{1,t} < \tfrac{1}{2} \le \hat{q}_{2,t} 
\right\} 
\cup 
\left\{ 
\hat{q}_{2,t} < \tfrac{1}{2} \le \hat{q}_{1,t} 
\right\}.
\]
Assume w.l.o.g. $q_1 < 1/2 < q_2$ and $q_1 + q_2 < 1$, for which $\bar{x}^\star = 0$. A mistake ($x_t=1$) occurs if:
\begin{enumerate}
    \item \textbf{False Agreement on 1:} $\hat{q}_{1,t} \ge 1/2$ and $\hat{q}_{2,t} \ge 1/2$. This implies $\hat{q}_{1,t} \ge 1/2$.
    \item \textbf{False Resolution:} Disagreement holds, but $\Sigma_t = \hat{q}_{1,t} + \hat{q}_{2,t} \ge 1$. (Recall that per Algorithm \ref{alg:binary}, the algorithm selects $x_t=1$ in a disagreement scenario only if $\Sigma_t \ge 1$). This implies $\Sigma_t \ge 1$.
\end{enumerate}
Thus, the mistake event is a subset of $\{ \hat{q}_{1,t} \ge 1/2 \} \cup \{ \Sigma_t \ge 1 \}$.
Let $\Delta_1 = 1/2 - q_1 > 0$ and $\Delta_\Sigma = 1 - (q_1+q_2) > 0$.

For the first term, we apply Hoeffding's inequality to the marginal $\hat{q}_{1,t}$.
For the second term, we define the sum variable $Z_s = b_s^1 + b_s^2$. Since the pairs $(b_s^1, b_s^2)$ are i.i.d. across time, the variables $Z_s$ are i.i.d. across time with range $[0, 2]$. The correlation between $b_s^1$ and $b_s^2$ affects the variance but not the independence or the range of $Z_s$.
Applying Hoeffding's inequality to the average $\Sigma_t = \frac{1}{t-1}\sum_{s=1}^{t-1} Z_s$:
\[
\mathbb{P}(\Sigma_t \ge 1) = \mathbb{P}(\Sigma_t - (q_1+q_2) \ge \Delta_\Sigma) \le e^{-\frac{2(t-1)\Delta_\Sigma^2}{(2-0)^2}} = e^{-(t-1)\Delta_\Sigma^2/2}.
\]
Combining these, we bound the probability of error:
\[
\mathbb{P}(x_t \neq \bar{x}^\star) \le e^{-2(t-1)\Delta_1^2} + e^{-(t-1)\Delta_\Sigma^2/2}.
\]

For all cases, $\mathbb{P}(x_t \neq \bar{x}^\star) \le \kappa e^{-c(t-1)}$ for constants $\kappa, c > 0$. Using \eqref{eq:binaryregretbound} and the bound \eqref{eq:termB} for Term (B), from \eqref{eq:regretdecompbinary} we get
\[
\bbE\{R^b_\cA(T)\} \le \sum_{t=1}^T \mathbb{P}(x_t \neq \bar{x}^\star) + 8Te^{-\delta^2 T / 2} \le \sum_{t=1}^\infty \left( 2e^{-c_1 t} + e^{-c_2 t} \right) + 8Te^{-\delta^2 T / 2} = O(1).
\]
\end{proof}

\section{Martingale Difference Inputs}
\label{app:martingale}

In this section, we extend our main result (Theorem \ref{thm:main} of Section \ref{sec:iid})
to the setting where the loss sequence satisfies a Martingale difference condition.

Let $\{\mathcal F_t\}_{t\ge 0}$ denote the natural filtration generated by the online
algorithm and the loss sequence. We assume that for every $x\in \cX$,
\begin{equation}
\label{eq:md_model}
\mathbb E\!\left\{\lambda_t(x)\mid \mathcal F_{t-1}\right\} = \mu(x),
\qquad
\lambda_t(x) := (f_t(x),g_t(x)),
\end{equation}
where $\mu(x)$ is a deterministic vector. We further assume that $f_t$ and $g_t$
satisfy Assumptions~\ref{cvx} and \ref{bddness} (boundedness, convexity, and Lipschitz continuity).

Recall the benchmark
\begin{equation}
\label{eq:COPT_md}
\bbE\{C_{\opt}\}
=
\bbE\left\{\min_{x\in \cX}
\max\!\left\{
\sum_{t=1}^T f_t(x),\;
\sum_{t=1}^T g_t(x)
\right\}\right\},
\end{equation}
and the deterministic surrogate benchmark
\begin{equation}
\label{eq:CbarOPT_md}
\bar C_{\opt}
=
T \min_{x\in \cX} \max_{\theta\in\Delta_2} \theta^\top \mu(x),
\end{equation}
and the expected {\it min-max} static regret is defined as before
$${\bar R}_\cA(T) = \sup_{\cD} \{ \bbE\{ C_\cA\} - {\bar C}_{\opt} \}.$$

\begin{lemma}[Benchmark Gap under Martingale Difference Inputs]
\label{lem:md_benchmark_gap}
Under the Martingale difference model~\eqref{eq:md_model} and
Assumptions~\ref{cvx} and \ref{bddness},
\[
\bbE\left\{\bigl|C_{\opt} - \bar C_{\opt}\bigr|\right\}
\;\le\;
O(\sqrt{T\log T}).
\]
\end{lemma}

\begin{proof}
The proof follows Appendix~\ref{app:regretconnection} verbatim with one modification.
For a fixed $x\in \cX$, the sequence
$\{\lambda_t(x)-\mu(x)\}_{t=1}^T$ is a bounded Martingale difference sequence.
Hence, the concentration step in Appendix~\ref{app:regretconnection},
which uses Hoeffding's inequality for i.i.d.\ variables, is replaced by
Freedman's inequality as follows.

Fix $x\in \cX$ and define
\[
Z_t(x) := \lambda_t(x) - \mu(x),
\qquad
\mathbb E\{Z_t(x)\mid\mathcal F_{t-1}\} = 0.
\]
For each coordinate $i\in\{1,2\}$, $|Z_{t,i}(x)| \le 2B$ almost surely, and define
\[
V_{T,i}(x)
:=
\sum_{t=1}^T
\mathbb E\!\left\{ Z_{t,i}(x)^2 \mid \mathcal F_{t-1} \right\}
\le 4B^2 T.
\]

We apply \emph{Freedman's inequality} \cite{freedman1975tail}, which states that for any
scalar Martingale difference sequence $\{Z_t\}$ with $|Z_t|\le b$ almost surely,
\[
\Pr\!\left(
\sum_{t=1}^T Z_t \ge u
\right)
\le
\exp\!\left(
-\frac{u^2}{2(V_T + bu/3)}
\right).
\]

This replaces the use of Hoeffding's inequality in Appendix~\ref{app:regretconnection}.
We apply Freedman's inequality \emph{coordinate-wise} to $\lambda_t(x)$, then similar to Appendix~\ref{app:regretconnection},
take a union bound over the two coordinates and an $\varepsilon$--net of $\cX$,
and then use Lipschitz continuity to extend the bound to all $x\in\cX$.
This yields
\[
\mathbb E\!\left\{
\sup_{x\in \cX}
\Big\|
\sum_{t=1}^T \lambda_t(x) - T\mu(x)
\Big\|_\infty
\right\}
=
O(\sqrt{T\log T}),
\]
which implies the stated bound exactly as in Appendix~\ref{app:regretconnection}.
\end{proof}

\begin{theorem}
\label{thm:mainMartingale}
Under Assumptions~\ref{cvx} and \ref{bddness}, for
$\eta_{\theta,t}\le \sqrt{\frac{2\ln 2}{B^2 t}}$ and
$\eta_{x,t} = \frac{D}{G\sqrt{t}}$,
the expected \emph{min--max} static regret
\eqref{defn:expregretintro} of Algorithm~\ref{alg:main}
under the Martingale difference model for $(f_t,g_t)$ satisfies
\[
\bar R_\cA(T) = O(\sqrt{T}).
\]
\end{theorem}

\begin{proof}
The regret decomposition and the bounds on $\bbE\{R_1\}$ and $\bbE\{R_2\}$
are identical to those in the proof of Theorem~\ref{thm:main}.
It remains to show that $\bbE\{R_3\}\le 0$.

Recall
\[
R_3
=
\min_{x\in \cX}\sum_{t=1}^T \Lambda_t(x,\theta_t)
-
\bar C_{\opt},
\qquad
\Lambda_t(x,\theta)=\theta^\top\lambda_t(x).
\]
Let $(x^\star,\theta^\star)$ solve
\[
\min_{x\in \cX}\max_{\theta\in\Delta_2}\theta^\top\mu(x).
\]
As in the proof of Lemma~\ref{lem:R3}, we have the deterministic inequality
\begin{equation}
\label{eq:R3_md_ub}
R_3
\le
\sum_{t=1}^T (\theta_t-\theta^\star)^\top \lambda_t(x^\star).
\end{equation}

Taking conditional expectation with respect to $\mathcal F_{t-1}$ and using
\eqref{eq:md_model},
\[
\mathbb E\!\left\{
(\theta_t-\theta^\star)^\top \lambda_t(x^\star)
\mid \mathcal F_{t-1}
\right\}
=
(\theta_t-\theta^\star)^\top \mu(x^\star).
\]
By optimality of $\theta^\star$, the right-hand side is non-positive almost surely.
Summing over $t$ and taking expectations in~\eqref{eq:R3_md_ub} yields
\[
\bbE\{R_3\} \le 0.
\]

\end{proof}

Combining Lemma~\ref{lem:md_benchmark_gap} and
Theorem~\ref{thm:mainMartingale}, the true {\it min-max} static regret of
Algorithm~\ref{alg:main} with respect to the benchmark
$\bbE\{C_{\opt}\}$ is $O(\sqrt{T\log T})$ under the Martingale difference condition.

\begin{algorithm}[H]
\caption{{\it Blocked Hedge}+OGD (for Markovian Inputs)}
\label{alg:blocked_main}
\begin{algorithmic}[1] 
\State \textbf{Input:} Block width $H$, step sizes $\eta_{x,m} > 0$, $\eta_{\theta,m} > 0$, feasible set $\mathcal{X}$
\State \textbf{Initialize:} $w_1 = (1,1)$, $\theta_1 = w_1 / \|w_1\|_1$, $x_1 \in \mathcal{X}$ arbitrary
\State Let $M = \lceil T/H \rceil$ be the number of blocks.

\For{$m=1,\dots,M$}
  \State \textbf{1. Freeze:} Set current action $x = x_m$ and weights $\theta = \theta_m$
  \State Set effective block length $H_m = \min(H,\, T-(m-1)H)$
  
  \State \textbf{2. Initialize block accumulators:}
  \State $\bar{g}_m = 0$, $\bar{\lambda}_m = 0$
  
  \State \textbf{3. Play and Observe (Block Loop):}
  \For{$t=(m-1)H+1, \dots, (m-1)H+H_m$}
    \State Play action $x_t = x$
    \State Observe losses $(f_t(x), g_t(x))$ and gradients $(\nabla f_t(x), \nabla g_t(x))$
    \State Evaluate loss vector $\lambda_t = (f_t(x), g_t(x))$
    \State Form surrogate gradient:
    \[
      \nabla \tilde{\Lambda}_t(x,\theta)
      := \theta_{1} \nabla f_t(x) + \theta_{2} \nabla g_t(x)
    \]
    \State \textit{Accumulate:}
    \State \quad $\bar{\lambda}_m \leftarrow \bar{\lambda}_m + \lambda_t$
    \State \quad $\bar{g}_m \leftarrow \bar{g}_m + \nabla \tilde{\Lambda}_t(x,\theta)$
 \EndFor
  
  \State \textbf{4. Block averaging:}
  \State $\bar{\lambda}_m \leftarrow \bar{\lambda}_m / H_m$
  \State $\bar{g}_m \leftarrow \bar{g}_m / H_m$

  \State \textbf{5. Updates (End of Block):}
  \State {\bf OGD:} $y_{m+1} = x_m - \eta_{x,m} \bar{g}_m$
  \State \quad\quad $x_{m+1} = \mathrm{Proj}_{\mathcal{X}}(y_{m+1})$
  \State {\it Hedge:} $w_{m+1,i} = w_{m,i} \exp(\eta_{\theta,m} \bar{\lambda}_{m,i})$, $i=1,2$
  \State Normalize: $\theta_{m+1} = w_{m+1}/\|w_{m+1}\|_1$
\EndFor
\end{algorithmic}
\end{algorithm}

\section{Markovian Inputs}
\label{app:markov}

In this section, we extend our main result (Theorem \ref{thm:main} of Section \ref{sec:iid})
to the setting where the loss sequence evolves according to a stationary Markov process.
Unlike the Martingale difference case, temporal dependence introduces a persistent bias
that requires modifying the algorithm to decouple the learning updates from the
instantaneous state of the Markov chain.

Let $\{S_t\}_{t\ge 1}$ be a time-homogeneous, ergodic Markov chain on a measurable
state space $\cS$ with transition kernel $P$ and stationary distribution $\pi$.
We assume the chain is \emph{uniformly ergodic}, i.e., there exist constants
$C>0$ and $\tau_{\mathrm{mix}}>0$ such that for any initial distribution $\nu$,
\[
\|\nu P^t-\pi\|_{\mathrm{TV}} \le C e^{-t/\tau_{\mathrm{mix}}}, \qquad \forall t\ge 1.
\]

Let $\lambda:\cX\times\cS\to\mathbb R^2$ be a measurable mapping and define
\[
\lambda_t(x) := \lambda(x,S_t) = (f_t(x),g_t(x)).
\]
We assume that $f_t$ and $g_t$ satisfy Assumptions~\ref{cvx} and \ref{bddness}.

Define the stationary mean
\begin{equation}
\label{eq:markov_mu}
\mu(x)
:=
\mathbb E_{\pi}[\lambda(x,S)],
\qquad \forall x\in\cX,
\end{equation}
which is independent of $t$ by stationarity.

Recall the benchmark
\begin{equation}
\label{eq:COPT_markov}
\bbE\{C_{\opt}\}
=
\bbE\!\left[
\min_{x\in \cX}
\max\!\left\{
\sum_{t=1}^T f_t(x),\;
\sum_{t=1}^T g_t(x)
\right\}
\right],
\end{equation}
and the deterministic surrogate benchmark
\begin{equation}
\label{eq:CbarOPT_markov}
\bar C_{\opt}
=
T \min_{x\in \cX} \max_{\theta\in\Delta_2} \theta^\top \mu(x),
\end{equation}
and the expected {\it min-max} static regret is defined as before
$$\bar R_\cA(T) = \sup_{P} \{ \bbE\{C_\cA\}- \bar C_{\opt}\}.$$

\begin{lemma}[Benchmark Gap under Markovian Inputs]
\label{lem:markov_benchmark_gap}
Under the Markovian model described above and Assumptions~\ref{cvx} and \ref{bddness},
\[
\bbE\left\{\bigl|C_{\opt} - \bar C_{\opt}\bigr|\right\}
\;\le\;
O(\sqrt{T\log T}) + O(\tau_{\mathrm{mix}}).
\]
\end{lemma}

\begin{proof}
The proof follows Appendix~\ref{app:regretconnection} with the sole modification that concentration of
$\sum_{t=1}^T \lambda_t(x)$ around $T\mu(x)$ uses Markov-chain concentration
instead of independence. Specifically, using uniform ergodicity and standard additive
functional concentration (e.g., \cite{paulin2015concentration}), for any fixed $x\in\cX$:
\[
\mathbb E\!\left\{
\Big\|
\sum_{t=1}^T \lambda_t(x) - T\mu(x)
\Big\|_\infty
\right\}
=
O(\sqrt{T\log T}) + O(\tau_{\mathrm{mix}}).
\]
All minimax and $\varepsilon$-net arguments of Appendix~\ref{app:regretconnection} remain unchanged, yielding the stated bound.
\end{proof}

Standard application of Algorithm~\ref{alg:main} fails in the Markovian setting because
the algorithm's updates $\theta_t$ become highly correlated with the instantaneous state
$S_t$, leading to an $O(T)$ regret.
To address this, we propose a \textbf{Blocked Hedge+OGD} strategy in Algorithm \ref{alg:blocked_main}.
The time horizon $T$ is divided into $M = \lceil T/H \rceil$ blocks of length $H$.
Within block $m$, the algorithm freezes its action $x_m$ and weights $\theta_m$,
accumulates losses, and performs a single update for both the {\it Hedge} and the OGD at the end of the block.

\begin{theorem}
\label{thm:mainMarkov}
Under Assumptions~\ref{cvx} and \ref{bddness}, running the Blocked Hedge+OGD
algorithm (Algorithm \ref{alg:blocked_main}) with block size $H =\Theta( T^{1/3}\tau_{\mathrm{mix}}^{2/3})$ yields an
expected {\it min-max} static regret of
\[
\bar R_\cA(T)
=
O(T^{2/3}\tau_{\mathrm{mix}}^{1/3}).
\]
\end{theorem}

\begin{proof}
Let $M = T/H$ be the number of blocks. We decompose the regret over the blocks similar to \eqref{defn:regretdecomp}.
The expected min-max static regret ${\bar R}_{\mathcal{A}}(T)$ is decomposed into three terms by adding and subtracting $\sum_{m=1}^M \Lambda^{(m)}(x_m, \theta_m)$ and $\min_{x \in \mathcal{X}} \sum_{m=1}^M \Lambda^{(m)}(x, \theta_m)$:$${\bar R}_{\mathcal{A}}(T) = \sup_{\mathcal{D}} \{ \mathbb{E}\{R_1\} + \mathbb{E}\{R_2\} + \mathbb{E}\{R_3\} \}$$
where:$$R_1= \max_{\theta \in \Delta_K} \sum_{m=1}^M \theta^{\top} \Lambda^{(m)}(x_m) - \sum_{m=1}^M \theta_m^{\top} \Lambda^{(m)}(x_m),$$ 
$$R_2 = \sum_{m=1}^M \theta_m^{\top} \Lambda^{(m)}(x_m) - \min_{x \in \mathcal{X}} \sum_{m=1}^M \theta_m^{\top} \Lambda^{(m)}(x),$$ and 
$$R_3=\min_{x \in \mathcal{X}} \sum_{m=1}^M \theta_m^{\top} \Lambda^{(m)}(x) - {\bar C}_\opt.$$

In Appendix \ref{app:blocked_proof}, we prove the following bounds
\begin{equation}\label{app:markov1}
  \mathbb E\{R_1 + R_2\} \le O(H \sqrt{M}) = O(H \sqrt{T/H}) = O(\sqrt{TH}).
\end{equation}
 \begin{equation}\label{app:markov2}
  \mathbb E\{R_3\} \le \sum_{m=1}^M O(\tau_{\mathrm{mix}}) = \frac{T}{H} O(\tau_{\mathrm{mix}}).
\end{equation}

Combining these, the expected {\it min-max} static regret is bounded by:
\[
\bar R_\cA(T) \le O(\sqrt{TH}) + O\left(\frac{T\tau_{\mathrm{mix}}}{H}\right).
\]
Minimizing this bound with respect to $H$ requires balancing $\sqrt{TH}$ with $T\tau_{\mathrm{mix}}/H$,
which yields the optimal block length $H =\Theta (T\tau_{\mathrm{mix}}^2)^{1/3}$.
Substituting this back gives the final bound $O(T^{2/3}\tau_{\mathrm{mix}}^{1/3})$.
\end{proof}

Combining Lemma~\ref{lem:markov_benchmark_gap} and
Theorem~\ref{thm:mainMarkov}, the true {\it min-max} static regret of the \textbf{Blocked Hedge+OGD} Algorithm \ref{alg:blocked_main}
with respect to the true benchmark $\bbE\{C_{\opt}\}$ is
\[
O(T^{2/3}\tau_{\mathrm{mix}}).
\]

\subsection{Proofs of \eqref{app:markov1} and \eqref{app:markov2}}\label{app:blocked_proof}

We provide the complete derivation for the upper bounds on the expected regret terms $\mathbb{E}\{R_1\}$, $\mathbb{E}\{R_2\}$, and 
$\mathbb{E}\{R_3\}$ for the \textbf{Blocked Hedge+OGD} algorithm with block size $H$. Let $M = \lceil T/H \rceil$ be the total number of blocks. We denote the blocks by $\mathcal{B}_1, \dots, \mathcal{B}_M$.\subsubsection{Upper Bound for $\mathbb{E}\{R_1\}$ }The term $R_1$ captures the regret of the weight updates $\theta_m$ against the best fixed distribution $\theta^\star$ over the sequence of blocks:$$R_1 = \sum_{m=1}^M (\theta_m - \theta^\star)^\top \Lambda^{(m)}(x_m),$$where $\Lambda^{(m)}(x) = \sum_{t \in \mathcal{B}_m} \lambda_t(x)$ is the cumulative loss vector for block $m$.\begin{proof}
\begin{enumerate}
\item The loss vector for each step $m$ is the block sum $\Lambda^{(m)}(x_m)$. Given that $\|\lambda_t\|_\infty \le B$ (Assumption 2), the magnitude of the loss vector is bounded by:$$\|\Lambda^{(m)}(x_m)\|_\infty \le \sum_{t \in \mathcal{B}_m} \|\lambda_t\|_\infty \le H \cdot B \text{.}$$\item  Although Algorithm \ref{alg:blocked_main} updates weights using the block-average loss $\bar{\lambda}_m = \Lambda^{(m)}/H_m$, this is mathematically equivalent to the gains version of Hedge running on the meta-sequence of blocks with an effective step size $\tilde{\eta}_{\theta,m} = \eta_{\theta,m}/H$.
\item  For a horizon of $M$ steps with losses bounded by $L_{\max} = HB$ and $K=2$ experts, the {\it Hedge} algorithm yields:
\[
R_1 \le HB \sqrt{2 M \ln 2} \text{.}
\]
\item Using $M \approx T/H$
\[
R_1 \le HB \sqrt{\frac{2T \ln 2}{H}} = B \sqrt{2TH \ln 2} \text{.}
\]
\end{enumerate}
Thus, $\mathbb{E}\{R_1\} = O(\sqrt{TH})$.
\end{proof}\subsubsection{Upper Bound for $\mathbb{E}\{R_2\}$}The term $R_2$ captures the regret of the action updates $x_m$ against the best fixed action $x^\star$:$$R_2 = \sum_{m=1}^M \left( \sum_{t \in \mathcal{B}_m} \theta_m^\top \lambda_t(x_m) - \sum_{t \in \mathcal{B}_m} \theta_m^\top \lambda_t(x^\star) \right) \text{.}$$\begin{proof}
\begin{enumerate}
\item  The effective gradient for block $m$ is the sum of gradients: $G^{(m)} = \sum_{t \in \mathcal{B}_m} \nabla (\theta_m^\top \lambda_t(x_m))$. Given $\|\nabla \lambda_t\|\le G$ (Assumption 3), the gradient norm is bounded by:$$\|G^{(m)}\| \le \sum_{t \in \mathcal{B}_m} \|\nabla \lambda_t\| \le H \cdot G \text{.}$$\item  Algorithm \ref{alg:blocked_main} performs OGD using the block-average gradient $\bar{g}_m = G^{(m)}/H_m$. In the regret analysis, this is equivalent to OGD using the cumulative gradient $G^{(m)}$ with an effective step size $\tilde{\eta}_{x,m} = \eta_{x,m}/H$.
\item For $M$ steps with gradient norms bounded by $\mathcal{G}_{\max} = HG$ and domain diameter $D$, OGD yields:
\[
R_2 \le D(HG) \sqrt{M} = DHG \sqrt{\frac{T}{H}} = DG \sqrt{TH} \text{.}
\]
\end{enumerate}
Thus, $\mathbb{E}\{R_2\} = O(\sqrt{TH})$.
\end{proof}

\subsubsection{Upper Bound for $\mathbb{E}\{R_3\}$} 
$$\begin{aligned}
R_3 &= \min_{x \in \mathcal{X}} \sum_{m=1}^M \theta_m^{\top} \Lambda^{(m)}(x) - {\bar C}_\opt, \\
&\stackrel{(a)}{\le} \sum_{m=1}^M \theta_m^{\top} \Lambda^{(m)}(x^\star) - \sum_{m=1}^M H (\theta^\star)^\top \mu(x^\star), \\
&\stackrel{(b)}{=} \sum_{m=1}^M \theta_m^{\top} (\Lambda^{(m)}(x^\star) - H \mu(x^\star)) + \sum_{m=1}^M (\theta_m - \theta^\star)^\top (H \mu(x^\star)), \\
&\stackrel{(c)}{=} \sum_{m=1}^M (\theta_m - \theta^\star)^\top (\Lambda^{(m)}(x^\star) - H \mu(x^\star)) + \sum_{m=1}^M (\theta_m - \theta^\star)^\top (H \mu(x^\star)) \\
& \quad + \sum_{m=1}^M (\theta^\star)^\top (\Lambda^{(m)}(x^\star) - H \mu(x^\star)), \\
&\stackrel{(d)}{\le} \sum_{m=1}^M \|\theta_m - \theta^\star\|_1 \|\Lambda^{(m)}(x^\star) - H \mu(x^\star)\|_\infty + \sum_{m=1}^M (\theta_m - \theta^\star)^\top (H \mu(x^\star))\\
& \quad  + \sum_{m=1}^M \|\theta^\star\|_1 \|\Lambda^{(m)}(x^\star) - H \mu(x^\star)\|_\infty, \\
&\stackrel{(e)}{\le} 3 \sum_{m=1}^M \|\Lambda^{(m)}(x^\star) - H \mu(x^\star)\|_\infty,
\end{aligned},$$
where $(a)$ follows from the suboptimality of the stationary optimal action $x^\star \in \mathcal{X}$, $(b)$ is obtained by adding and subtracting the mixed term $\sum_{m=1}^M \theta_m^\top (H \mu(x^\star))$, $(c)$ uses the algebraic identity $\theta_m^\top V = (\theta_m - \theta^\star)^\top V + (\theta^\star)^\top V$, where $V = \Lambda^{(m)}(x^\star) - H \mu(x^\star)$, to center the algorithm's weights around the stationary optimal weights $\theta^\star$, $(d)$ follows from H{o}lder's inequality, where $u^\top v \le \|u\|_1 \|v\|_\infty$, and finally $(e)$ utilizes three key properties:
\begin{enumerate}
\item since $\theta_m, \theta^\star \in \Delta_K$, we have $\|\theta_m - \theta^\star\|_1 \le \|\theta_m\|_1 + \|\theta^\star\|_1 = 2$, \item stationary optimality of $\theta^\star$ ensures $(\theta_m - \theta^\star)^\top \mu(x^\star) \le 0$ for all $\theta_m \in \Delta_K$, so $ \sum_{m=1}^M (\theta_m - \theta^\star)^\top (H \mu(x^\star))$ is bounded by zero, and 
\item  $\|\theta^\star\|_1 = 1$ because it is a probability distribution on the simplex.
 \end{enumerate}
To complete the proof, we now bound the expected deviation term $\|\Lambda^{(m)}(x^\star) - H\mu(x^\star)\|_\infty$ using the mixing properties of the chain. By the law of iterated expectations, we have $\mathbb{E}[\|\Lambda^{(m)}(x^\star) - H\mu(x^\star)\|_\infty] = \mathbb{E}[\mathbb{E}[\|\sum_{t \in \mathcal{B}_m} (\lambda_t(x^\star) - \mu(x^\star))\|_\infty \mid \mathcal{F}_{\text{start}(m)}]]$, where $\mathcal{F}_{\text{start}(m)}$ is the filtration at the start of block $m$.
 
 \begin{enumerate}
\item For a uniformly ergodic Markov chain, the expected deviation of the state from stationarity decays exponentially with time. Let $S_{\text{start}(m)}$ be the state at the start of block $m$. For the $k$-th step within the block ($1 \le k \le H$): $$\left\|\mathbb{E}[\lambda_{(m-1)H+k}(x^\star) \mid S_{\text{start}(m)}] - \mu(x^\star) \right\|_\infty \le C e^{-k/\tau_{\mathrm{mix}}} $$\item We sum this deviation over the $H$ steps in the block:
\[
\left\| \mathbb{E} \left[ \sum_{t \in \mathcal{B}_m} (\lambda_t(x^\star) - \mu(x^\star)) \;\Bigg|\; \mathcal{F}_{\text{start}(m)} \right] \right\|_\infty \le \sum_{k=1}^H C e^{-k/\tau_{\mathrm{mix}}}.
\]
This geometric series is bounded by a constant proportional to the mixing time
\[
\sum_{k=1}^\infty C e^{-k/\tau_{\mathrm{mix}}} = O(\tau_{\mathrm{mix}})
\]
Thus, the total bias accumulated \textbf{per block} is $O(\tau_{\mathrm{mix}})$.
\item  Summing this constant error over the $M$ blocks
\[
\mathbb{E}\{R_3\} \le 3 \sum_{m=1}^M O(\tau_{\mathrm{mix}}) = M \cdot O(\tau_{\mathrm{mix}}) = \frac{T}{H} O(\tau_{\mathrm{mix}}) 
\]
\end{enumerate}
Thus, we get $$\mathbb{E}\{R_3\} = O\left( \frac{T \tau_{\mathrm{mix}}}{H} \right).$$

\end{document}